%% file: main.tex
\pdfoutput=1
\documentclass[10pt]{article}
\usepackage[preprint]{tmlr}

\input{math_commands.tex}

\usepackage{hyperref}
\usepackage{url}
\usepackage{graphicx}
\usepackage{booktabs}
\usepackage{multirow}
\usepackage{amsmath,amssymb}
\usepackage{xcolor}
\usepackage[capitalise,noabbrev]{cleveref}
\usepackage{enumitem}
\setlist{nosep,leftmargin=1.4em}

\newcommand{\Rc}{R}
\newcommand{\enc}{\phi}
\newcommand{\dv}{D\'ej\`a VAE}
\newcommand{\dvI}{D\'ej\`a VAE-I}
\newcommand{\dvP}{D\'ej\`a VAE-P}

\newcommand{\ci}[2]{{\small[#1,\,#2]}}

\title{D\'ej\`a VAE: When Do Autoencoder Traces Reveal Generated Images?}

\author{\name Saugat Pandey \email spandey2@furman.edu \\
      \addr Department of Computer Science\\
      Furman University}

\begin{document}
\maketitle

\begin{abstract}
A latent diffusion model's output has already passed through its autoencoder, so a second round trip changes it very little.
This ``d\'ej\`a vu'' underlies reconstruction-based detectors of generated images; we analyse when it reveals a generated image and when it does not.
Read by a frozen vision foundation model without training, the trace is strongest in intermediate layers, and, unlike pixel-level reconstruction error, it treats blur as a sign of a real image.
It belongs to the generator's exact decoder: on Synthbuster, the exact decoders of Stable Diffusion 1.3, 1.4, 2 and XL are the strongest of eight autoencoders and catch 94--100\% of their generators' images at a 5\% false-alarm rate (SDXL's version identified post hoc), while near-identical versions catch 7--42\%.
A detector trained only on real photographs and their autoencoder copies matches one trained on 162k generated images on GenImage (+2.3 points, 95\% CI [0.9, 3.7]) and complements it in fusion, but one resize with JPEG re-save, or a mild blur, costs it 19--27 points.
Two lessons concern benchmarks. The in-the-wild benchmark Chameleon can be solved from image width alone. And on Synthbuster, how the real photographs are matched to the generated images' size decides whether our detector leads or trails ($+4.4$ or $-9.7$ points): the matching protocol alone can reverse a detector comparison.
\end{abstract}

\input{sections/introduction}
\input{sections/related}
\input{sections/setup}
\input{sections/where}
\input{sections/decoder}
\input{sections/zerofake}
\input{sections/synthbuster}
\input{sections/wild}
\input{sections/chameleon}
\input{sections/fusion}
\input{sections/limitations}

\subsubsection*{Broader Impact Statement}
Detectors of generated images are used in journalism, content moderation and forensics, where a false accusation can harm a real person. We report false-alarm rates on degraded real photographs for every method, and we show that a widely used benchmark overstates in-the-wild performance. Our results also show how this kind of detector can be evaded: one resize and JPEG re-save, or a mild blur, costs the trained detectors 19--27 points. We report this because the weakness exists whether or not it is published, and because defenders who deploy such detectors need to know it, for example to combine them with detectors that are robust to processing, as our fusion results suggest. We release code and per-image scores, not images, and we informed the Chameleon authors of the metadata shortcut before releasing this work.

\bibliography{main}
\bibliographystyle{tmlr}

\appendix
\input{sections/appendix}

\end{document}

%% file: math_commands.tex
\usepackage{amsmath,amsfonts,bm}

\def\eqref#1{equation~\ref{#1}}

\def\1{\bm{1}}

\DeclareMathAlphabet{\mathsfit}{\encodingdefault}{\sfdefault}{m}{sl}
\SetMathAlphabet{\mathsfit}{bold}{\encodingdefault}{\sfdefault}{bx}{n}

%% file: sections/introduction.tex
\section{Introduction}
\label{sec:intro}

\begin{figure}[t]
    \centering
    \includegraphics[width=\linewidth]{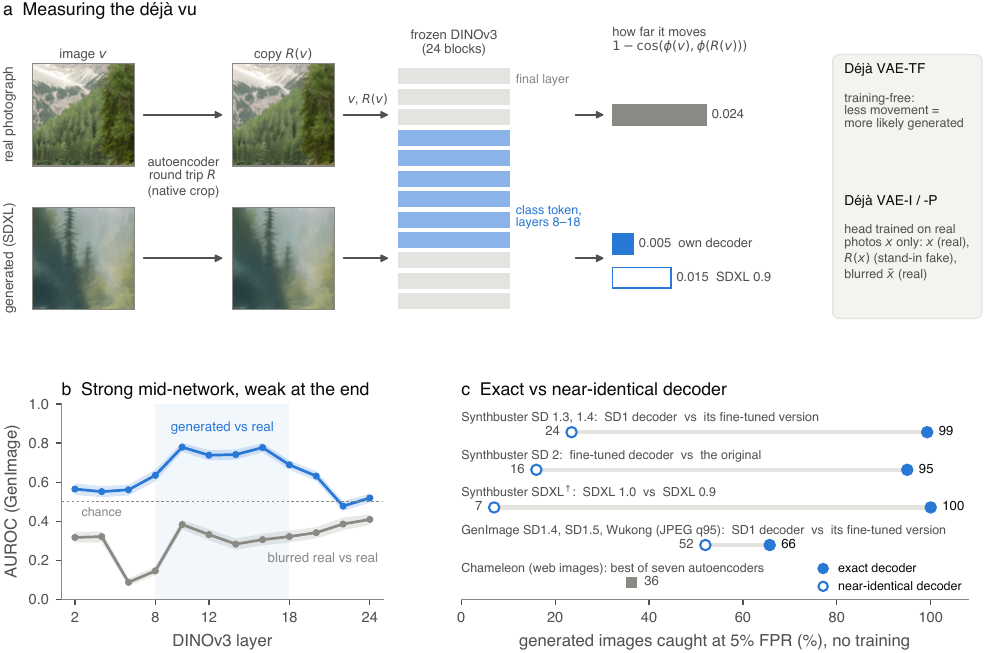}
    \caption{\textbf{What we measure, and the two findings that frame the paper.} \textbf{(a)} The instrument, on real data: a Synthbuster \citep{bammey2024synthbuster} SDXL image (median of its class by fixed-point score) and the RAISE \citep{dangnguyen2015raise} photograph it was prompted from (68th percentile among real images). An autoencoder round trip $\Rc$ at native resolution returns a copy that looks identical to each image. A frozen DINOv3 reads image and copy, and the bars show how far the class token moves (cosine distance, mean over layers 8--18). The generated image barely moves under its own decoder (SDXL~1.0), and three times as far under the 0.9 version of the same autoencoder. \dv-TF scores this movement without training; \dvI{} and \dvP{} are small heads trained on real photographs only, with their copies as stand-in fakes and blurred photographs as counterfactual reals (\cref{sec:setup}). \textbf{(b)} Without training, the signal is strongest in intermediate layers. AUROC of the training-free signal per DINOv3 layer on GenImage (blue; mean over eight generators), and of blurred real images against the originals (grey; a high value would mean that the score mistakes blur for generation). Bands: 95\% bootstrap intervals (\cref{sec:where}). \textbf{(c)} It belongs to the exact decoder. Share of generated images caught at a 5\% false-positive rate by the training-free score with the generator's own decoder (filled) and with a near-identical version of it (open). Synthbuster images are compared as stored (lossless) against real photographs downscaled to their size; with cropped reals the open markers move to 16.5\%, 42.0\% and 30.5\%, and the exact decoder remains the strongest of eight autoencoders. GenImage images are first re-saved as JPEG q95, and the effect is weaker there (\cref{sec:decoder}). $^\dagger$The matching SDXL version was identified post hoc. On web images (Chameleon), no autoencoder in our bank restores the signal.}
    \label{fig:overview}
\end{figure}

Most images produced by today's generators are decoded by an autoencoder.
A latent diffusion model denoises in a compressed latent space and then decodes the result into pixels \citep{rombach2022ldm}, so its output lies, by construction, close to the set of images that the decoder can produce.
Pass such an image through the same autoencoder once more and it barely moves. A real photograph moves further, because it never had to fit the decoder.
This asymmetry, an image being a near fixed point of an autoencoder, is the premise behind reconstruction-based detectors \citep{ricker2024aeroblade} and behind training on reconstructed real images \citep{rajan2025aligned}.
We call it the \emph{d\'ej\`a vu} of a generated image (\cref{fig:overview}).

The premise is appealing because it promises detection without examples of the generator.
If generated images are recognisable by how little they change under an autoencoder, a detector could be trained on real photographs alone and still catch every generator that uses that autoencoder.
Existing detectors only partly test this promise. Reconstruction-error detectors use the signal without training, but they are fragile to compression and flag degraded real photographs (\cref{sec:zerofake}). Detectors trained with reconstructions of real images as fakes \citep{rajan2025aligned} train a full network end to end, so where the trace is visible in a representation, and what destroys it, is not isolated.

This paper is an analysis of that premise, not a new detector. We measure the d\'ej\`a vu with a frozen vision foundation model, DINOv3 \citep{simeoni2025dinov3}, and build a measuring instrument around it, \dv{} (\cref{fig:overview}a). \dv{} copies an image with an autoencoder and reads image and copy at intermediate layers. It then either scores how little the image moved (\dv-TF, no training) or applies a small head trained on real photographs and their copies only. \dvI{} reads the image alone, and \dvP{} reads it together with its copy. Training without generated images is not new \citep{rajan2025aligned,cozzolino2024zed,karageorgiou2025spai}; we use it because it isolates the one trace we want to study. Holding backbone, head and protocol fixed, we then vary one factor at a time: the layer, the backbone, how the copy is made, the decoder, the processing after generation, and the benchmark.
With this instrument we ask six questions, each answered in its own section:
\begin{enumerate}[nosep,leftmargin=*,topsep=2pt]
    \item \textbf{Where is the trace visible?} Without training, mainly in intermediate layers and much less at the final one; there, unlike pixel-level reconstruction error, it treats blur as a sign of a real image (\cref{sec:where}). In our setup, probes trained on generated images flag only 15--25\% of autoencoder copies of real photographs, reproducing the blindness reported for final-layer probes \citep{zhou2026simplicity}; heads trained on the copies flag 50\% of them at the final layer and 98\% at layers 8--18 (\cref{app:ablation}).
    \item \textbf{Whose trace is it?} The generator's exact decoder. On Synthbuster, without training, the exact decoder is the strongest of eight autoencoders for each of SD~1.3, 1.4, 2 and XL (the SDXL version identified post hoc) and catches 94--100\% of their images, while near-identical versions of the same decoders catch 7--42\%, however the real photographs are matched in size. Retraining \dv{} with a different copy decoder moves results in the directions we predicted (\cref{sec:decoder}).
    \item \textbf{Does a detector trained without fakes work?} On GenImage, yes: \dvI{} detects 87.5\% of fakes at a 5\% false-alarm rate, on par with or above a detector trained on 162k generated images, and far above two training-free baselines (\cref{sec:zerofake}). On Synthbuster, whether it leads or trails that detector depends on how the real photographs are matched to the generated images' size ($+4.4$ or $-9.7$ points): the matching alone can reverse a detector comparison (\cref{sec:synthbuster}).
    \item \textbf{What breaks it?} One resize with JPEG re-save, or a mild blur, costs the trained detectors 19--27 points, while a semantic detector loses at most 8; processing-aware training does not help on web images (\cref{sec:wild}).
    \item \textbf{Is the in-the-wild benchmark measuring what we think?} Every real image in Chameleon \citep{yan2025sanity} is exactly 500 pixels wide, while only 1 of 11{,}170 generated images is. We build a size-matched version and separate the shortcut from the cost of processing (\cref{sec:chameleon}).
    \item \textbf{Is the signal still useful?} As a complement: fusing \dv{} with a semantic detector improves it on all three benchmarks, beats control ensembles, and keeps false alarms on degraded photographs at the base rate (\cref{sec:fusion}).
\end{enumerate}
Along the way we contribute a size-matched Chameleon with a processing-only control, two matched versions of Synthbuster, and per-image scores for every method, all under one protocol: thresholds from real images only, paired bootstrap intervals, and predictions written down before the experiments that test them (\cref{sec:protocol}). Code and scores are available at \url{https://github.com/iamsaugatpandey/deja-vae}.

We do not propose a state-of-the-art in-the-wild detector. Our claim is narrower and, we argue, more useful: we pin down when autoencoder traces reveal generated images, why they stop doing so on the open web, and how a benchmark can hide the difference.

%% file: sections/related.tex
\section{Related work}
\label{sec:related}

\paragraph{Detectors built on frozen foundation models.}
Linear probes and small heads on frozen CLIP or DINO features generalise across generators better than detectors trained end to end \citep{ojha2023universal}. SimLBR \citep{dhakal2026simlbr} trains a small head on DINOv3 class-token features and is the strongest such detector in our setting. \citet{zhou2026simplicity} show that linear probes on the pooled final-layer features of recent foundation models, DINOv3 among them, match specialised detectors on standard benchmarks, but are ``essentially blind'' to real images that have only been passed through a diffusion model's autoencoder. RIGID \citep{he2024rigid} is training-free: it flags an image whose DINOv2 embedding changes a lot under small Gaussian noise. All of these read the backbone's final layer. Without training, we find the autoencoder trace mainly in intermediate layers and much weaker at the final layer (\cref{sec:where}). In our setup, probes trained on generated images, like those of \citet{zhou2026simplicity}, also flag only 15--25\% of autoencoder copies of real photographs. Heads trained on the copies flag 50\% of them at the final layer and 98\% at layers 8--18 (\cref{app:ablation}), so the blindness reflects both what a probe is trained on and which layer it reads.

\paragraph{Reconstruction and the autoencoder fixed point.}
DIRE \citep{wang2023dire} and LaRE$^2$ \citep{luo2024lare} detect diffusion images by how well a diffusion model reconstructs them, in pixel space and in latent space. AEROBLADE \citep{ricker2024aeroblade} scores an image by its LPIPS distance to its own autoencoder reconstruction, taking the minimum over several autoencoders; it needs no training, but it requires the generator's autoencoder. AEDR \citep{aedr2026} uses the ratio of two successive autoencoder reconstruction errors for training-free attribution. A second line uses reconstructions of real images as training data. DRCT \citep{chen2024drct} treats diffusion reconstructions of real images as hard negatives for contrastive training. AlignedForensics \citep{rajan2025aligned} trains a detector end to end with autoencoder reconstructions of real images as the fakes, so that real and fake differ only in the autoencoder trace. DDA \citep{chen2025dda} additionally aligns the high frequencies and mixes pixels between real images and their reconstructions. B-Free \citep{guillaro2025bfree} creates training fakes from real images with a diffusion model's inpainting. Like AlignedForensics and DDA, our detector uses reconstructions of real images instead of generated images. Our aim differs: we use such a detector as an instrument to ask where the trace is visible, what it requires and what destroys it, holding the backbone, head and protocol fixed. Methodologically, ours reads a frozen foundation model at intermediate layers instead of training a network, copies at native resolution, optionally compares each image with its copy, and adds blur-matched counterfactual real images, without which false alarms on degraded photographs rise to 13--40\% (\cref{sec:zerofake}). We include an AlignedForensics-style reconstruction probe on the same frozen features as a baseline.

\paragraph{Detection without generated images.}
Training without generated images is not new. ZED \citep{cozzolino2024zed} measures how surprising an image is to a lossless coder trained on real images. SPAI \citep{karageorgiou2025spai} learns the spectral distribution of real images through a reconstruction pretext task and flags images that do not fit it. Forensic self-descriptions \citep{nguyen2025selfdesc} model the residuals of real images and use them for zero-shot detection, attribution and clustering. \citet{brokman2025manifold} derive zero-shot scores from the probability manifold of a pre-trained diffusion model. These methods model real images in general; ours targets one specific trace, the autoencoder fixed point, which lets us test when that trace is present. We compare only with detectors that read the same frozen backbone or rely on the same autoencoder signal, so that differences can be attributed to the signal rather than to backbones and training data (\cref{sec:limitations}).

\paragraph{Benchmarks and shortcuts.}
Shortcut learning, where a model succeeds for the wrong reason, is well documented \citep{geirhos2020shortcut}. For generated images, \citet{grommelt2024fakejpeg} showed that GenImage's real images are JPEG-compressed while its fakes are PNG, and that its fakes cluster in a few sizes, so detectors can learn compression and size instead of generation. \citet{zhou2026fragile} show that conclusions about training-free detectors on GenImage flip with the LPIPS backbone, the preprocessing and the score direction, and that RIGID's score is inverted on some latent diffusion generators; our RIGID results agree. Chameleon \citep{yan2025sanity} was introduced as a harder, in-the-wild benchmark and is now used to evaluate detectors on frozen foundation models \citep{zhou2026simplicity}; we find that it has its own, stronger size shortcut (\cref{sec:chameleon}). Synthbuster \citep{bammey2024synthbuster} pairs each generated image with a real photograph, which lets us remove size as a cue image by image, and shows that how this is done matters (\cref{sec:synthbuster}).

%% file: sections/setup.tex
\section{Measuring the d\'ej\`a vu}
\label{sec:setup}

\subsection{The autoencoder copy and the fixed-point signal}
\label{sec:signal}

Let $\Rc(\cdot)$ denote an autoencoder round trip: encode an image, take the mode of the latent distribution, and decode. For an image $x$, we call $\Rc(x)$ its \emph{autoencoder copy}. Unless stated otherwise, $\Rc$ is the Stable Diffusion autoencoder with the fine-tuned (ft-MSE) decoder, the checkpoint that ships with Stable Diffusion~2 \citep{rombach2022ldm}.\footnote{This checkpoint shares its encoder with Stable Diffusion~1.x, whose original decoder it replaces. We use the original decoder separately in \cref{sec:decoder}.}
The copy is made at native resolution. We take a centre crop of at most $256\times256$ pixels, with its offset and size on the autoencoder's 8-pixel grid, and copy that crop without resizing it first.

Let $\enc_\ell(v)\in\mathbb{R}^{1024}$ be the class token of a frozen backbone at layer $\ell$ for an image $v$, resized to the backbone's input size. Our main backbone is DINOv3 ViT-L/16 \citep{simeoni2025dinov3}; we replicate the measurements with DINOv2 ViT-L/14 \citep{oquab2024dinov2}.
The \emph{fixed-point signal} of $v$ at layer $\ell$ is how little its representation moves under the copy:
\begin{equation}
    s_\ell(v) \;=\; -\log\!\big(1-\cos\big(\enc_\ell(v),\, \enc_\ell(\Rc(v))\big)\big).
    \label{eq:signal}
\end{equation}
A higher $s_\ell$ means $v$ is closer to a fixed point of $\Rc$, and more likely generated. Throughout, the \emph{d\'ej\`a vu} is the phenomenon, the \emph{trace} is what a decoder leaves in its output that makes the phenomenon possible, and the \emph{signal} is what we measure, $s_\ell$ or a head trained on it. \Cref{eq:signal} at layer 16 is our training-free detector, \dv-TF.

\paragraph{How the layers were chosen.} The layer, and the layer set used by the trained heads below, were chosen from a training-free pilot diagnostic on 40 generated images per GenImage generator and 150 Chameleon fakes, drawn from the evaluation pools. The heads of the study's first version read the final-layer class token, and that version had already been scored on the full GenImage and Chameleon test sets when the diagnostic led us to layers 8--18. The switch raised GenImage and lowered Chameleon, and we kept it. GenImage and Chameleon therefore informed the layer choice; Synthbuster, added later, is the only benchmark held out from it, and there the choice does not transfer cleanly (\cref{sec:finallayer}).

\subsection{A detector that never sees a generated image}
\label{sec:detector}

\paragraph{Training rows.}
We start from $n=5{,}000$ real training photographs (the ImageNet images in GenImage's training split). From each real image $x$ we build three training rows. Each row is an image $u$ together with its own copy $\Rc(u)$:
\begin{itemize}[nosep,leftmargin=*,topsep=2pt]
    \item \textbf{real}: $u=x$, label 0;
    \item \textbf{simulated fake}: $u=\Rc(x)$, label 1. This is an image that has already been through the autoencoder once, as a latent diffusion output has;
    \item \textbf{counterfactual real}: $u=\tilde x$, label 0, where $\tilde x$ is $x$ blurred with a Gaussian whose width is chosen so that $\mathrm{PSNR}(x,\tilde x)=\mathrm{PSNR}(x,\Rc(x))$.
\end{itemize}
The counterfactual row matters. A copy $\Rc(x)$ is smoother than $x$, so without it a detector can learn that ``smooth means fake'' and flag every soft or slightly blurred real photograph. The counterfactual has the same loss of detail as the copy but has not been through the autoencoder. All three rows of a real image share one JPEG re-save with quality drawn uniformly from 70 to 100, so that compression carries no label information.

\paragraph{Heads.}
Let $a=\enc(u)$ and $b=\enc(\Rc(u))$, where $\enc$ concatenates the DINOv3 class tokens of layers $\{8,10,12,14,16,18\}$ (6{,}144 dimensions). We train two heads with the same architecture, a three-layer perceptron (input $\to$ 512 $\to$ 256 $\to$ 1) as in SimLBR:
\begin{itemize}[nosep,leftmargin=*,topsep=2pt]
    \item \textbf{\dvI} (image only) reads $a$;
    \item \textbf{\dvP} (paired) reads $[a,\; a-b,\; a\odot b]$, the image together with how it moved under the copy.
\end{itemize}
Both are trained for 2{,}000 steps (batch 500, Adam, learning rate $10^{-4}$, weight decay $10^{-2}$, dropout 0.5, class-balanced loss) with three seeds. At test time an image $v$ is scored on its own view and its copy, exactly as in training.

\subsection{Baselines}
\label{sec:baselines}

All learned baselines use the same frozen DINOv3 backbone and the same head architecture. The two training-free baselines use their published scoring rules.
\begin{itemize}[nosep,leftmargin=*,topsep=2pt]
    \item \textbf{SimLBR (released)} \citep{dhakal2026simlbr}: the authors' head, trained on 162k real and 162k Stable Diffusion~1.4 images from GenImage, on final-layer class tokens of images squashed to $256\times256$. Training fakes are replaced by feature blends $\alpha\,\enc(x_{\text{real}})+(1-\alpha)\,\enc(x_{\text{fake}})$ with $\alpha\sim\mathcal U(0.5,0.8)$.
    \item \textbf{SimLBR (retrained)}: the same recipe retrained on 10k real and 10k SD~1.4 images, to separate the recipe from the training scale.
    \item \textbf{Supervised probe}: the same head trained on real versus SD~1.4 images without blending.
    \item \textbf{Reconstruction probe}: an AlignedForensics-style head \citep{rajan2025aligned} trained on real images versus their autoencoder reconstructions, on final-layer features of squashed images. Like \dv, it uses no generated images.
    \item \textbf{RIGID} \citep{he2024rigid}: the cosine similarity between the embeddings of an image and of the image plus Gaussian noise ($\lambda=0.05$). We report DINOv2 ViT-L/14 as published, both noise conventions (the paper does not specify one; \cref{app:rigid}), and DINOv3 for backbone parity.
    \item \textbf{AEROBLADE} \citep{ricker2024aeroblade}: the minimum, over the SD1, SD~2 and Kandinsky~2.1 autoencoders, of the LPIPS distance at the second VGG layer between an image and its reconstruction. It uses the same native-resolution crop as \dv.
\end{itemize}

\subsection{Evaluation protocol}
\label{sec:protocol}

\paragraph{Benchmarks.}
\emph{GenImage} \citep{zhu2023genimage}: eight generators (ADM, BigGAN, GLIDE, Midjourney, VQDM, Stable Diffusion~1.4 and 1.5, Wukong), with 2{,}000 test fakes each, 11{,}200 test reals and 4{,}800 held-out calibration reals. GenImage's real images are JPEG files and its fakes are PNG files \citep{grommelt2024fakejpeg}. We therefore re-save every evaluation image as JPEG at quality 95, so that the file format carries no label information.
\emph{Chameleon} \citep{yan2025sanity}: 14{,}863 real and 11{,}170 generated images collected from the web. We hold out 4{,}458 of the reals for calibration.
\emph{Size-matched Chameleon} and its \emph{processing-only control} are defined in \cref{sec:chameleon}.
\emph{Synthbuster} \citep{bammey2024synthbuster}: 1{,}000 images from each of nine recent generators, each generated from a prompt describing one photograph of RAISE-1k \citep{dangnguyen2015raise} and named after it. The generators output fixed sizes from $256\times256$ (GLIDE) to $2688\times1536$ (Firefly), while the RAISE photographs are camera images about 4{,}300--4{,}900 pixels wide, so size alone would separate the classes. We therefore pair each generated image with a real image made from its own RAISE photograph at exactly its size, in two ways. \emph{Downscaled} (pre-specified): the photograph is centre-cropped to the generated image's aspect ratio and resized to its size (bicubic, antialiased; median factor 1.6 for Firefly to 12.8 for GLIDE). \emph{Cropped} (added after the downscaled results): a native-resolution centre crop at the generated image's size, with no resampling. Neither is neutral. Downscaling could itself separate the classes, and a crop shows a small part of the scene at the camera's pixel scale; \dv{}'s view of a cropped real covers about 5\% of the photograph's width, and the nine cropped reals of one photograph give \dv{} the same view. The generated images are left untouched; according to the dataset's documentation, none was compressed or resampled. We split the RAISE photographs 50/50 (seed 23) into a calibration and a test half. One photograph is corrupt on the dataset's server and is excluded with its nine generated images, leaving 4{,}500 calibration reals, 4{,}491 test reals and 8{,}991 generated images under each matching. Like GenImage, every Synthbuster image is re-saved as JPEG q95 before scoring.

\paragraph{Thresholds from real images only.}
For each method we set the decision threshold by split conformal calibration on the calibration reals \citep{vovk2005algorithmic,angelopoulos2023conformal}, targeting a false-positive rate of $\varepsilon\in\{5\%,1\%\}$. No generated image is used to set a threshold or a fusion weight. Generated images influenced design choices only through the choice of layers (\cref{sec:signal}). This mirrors deployment, where the detector's operator knows what real images look like but not which generator comes next.

\paragraph{Metrics.}
The primary metric is the true-positive rate at 5\% false positives (TPR@5\%), per generator, averaged over generators (\emph{mean}) and at the worst generator (\emph{worst}). We also report TPR@1\%, AUROC, and the false-positive rate on \emph{degraded real images}: test reals blurred with $\sigma\sim\mathcal U(0.5,1.5)$ and re-saved like the other test images. A detector that flags these is penalising image quality, not generation.

\paragraph{Uncertainty and comparisons.}
Every comparison is paired: the methods score the same images and use thresholds from the same calibration reals. For each difference we report the mean of its paired-bootstrap distribution ($B=1{,}000$) and the 95\% percentile interval (the mean can differ by 0.1 points from the difference of table entries). Each draw resamples the calibration reals and each generator's test fakes once, and every method and seed is evaluated on that same draw; on Synthbuster a draw resamples RAISE photographs instead, because each contributes nine matched reals and up to nine generated images. We call a difference significant only when its 95\% interval excludes zero. The intervals reflect the sampling of images; training-seed variation is averaged out. Across the three seeds, the standard deviation of mean TPR@5\% is at most 0.9 points on GenImage and at most 2.8 points on the other benchmarks for every learned method, and \dvI{} is above the released SimLBR head for each seed separately on GenImage and on Synthbuster with downscaled reals. Each interval covers one comparison; we do not correct for multiplicity, and for each question we report every comparison we ran, not only the significant ones.

\paragraph{Pre-specified predictions.}
For the experiments added late in the study (Synthbuster, its decoder bank, the CLIP layer curve and retraining with a different decoder), we wrote down the protocol and the expected outcomes before running them, and we report every prediction with its outcome, including those that failed. Analyses added after seeing a result are labelled \emph{post hoc}.

\paragraph{Fusion.}
To combine two detectors we standardise each one's scores by the mean and standard deviation of its scores on the calibration reals and add them. No weight is tuned.

%% file: sections/where.tex
\section{In which layers is the d\'ej\`a vu visible?}
\label{sec:where}

\textit{Question: in which layers of a frozen backbone, and under which copy conditions, does a generated image move less than a real one?}

\paragraph{Setup.}
For 400 real GenImage images, 200 images from each of the eight generators, and 300 real and 300 generated Chameleon images, we compute the fixed-point signal $s_\ell$ (\cref{eq:signal}) at every second layer of DINOv3 and DINOv2, and, as a backbone trained on image--text pairs rather than self-supervised, of CLIP ViT-L/14 \citep{radford2021clip}, together with two alternatives: the mean distance over patch tokens, and the pixel-level PSNR between $v$ and $\Rc(v)$. We report the AUROC of real against generated per generator, with bootstrap intervals over images. These diagnostic sets are drawn from the evaluation pools and are smaller than the full evaluation of \cref{sec:zerofake}; the numbers in this section and in the decoder banks (\cref{sec:decoder}) come from them. Real images also receive the two training-row measurements: a simulated fake $\Rc(x)$ against $x$, and a blur-matched counterfactual against $x$.

\begin{figure}[t]
    \centering
    \includegraphics[width=0.85\linewidth]{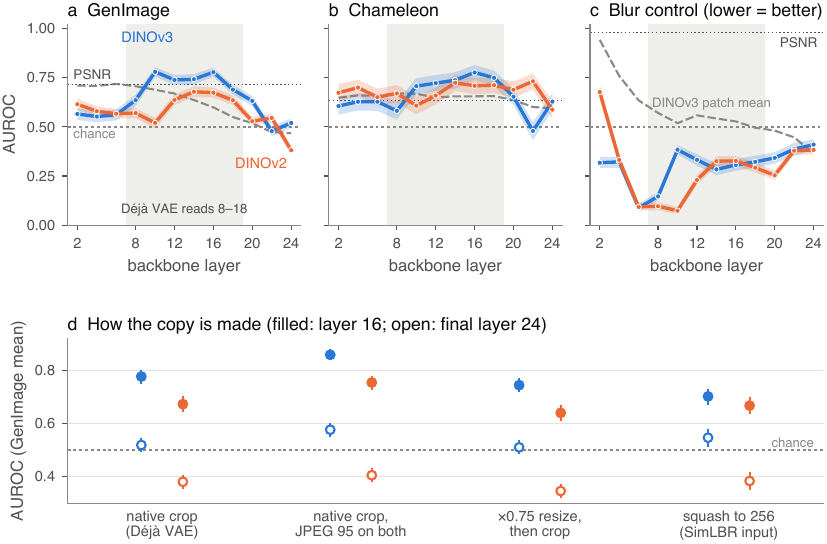}
    \caption{\textbf{The d\'ej\`a vu is visible in intermediate layers, and it is not a smoothness detector.} AUROC of the training-free signal $-d_\ell(v, \Rc(v))$ (class token) at every second layer of DINOv3 (blue) and DINOv2 (orange). \textbf{(a)} GenImage, mean over the eight generators (200 fakes each, 400 reals). \textbf{(b)} Chameleon (300 + 300). \textbf{(c)} Blur control: real images blurred to the PSNR of their own autoencoder copy, scored against the unblurred reals; a high value means the score mistakes smoothness for generation. Grey dotted: PSNR$(v, \Rc(v))$; grey dashed: DINOv3 mean patch-token distance. Shaded columns: the layers \dv{} reads. \textbf{(d)} How the copy is made, at layer 16 (filled) and the final layer (open). Bands and bars: 95\% bootstrap intervals (reals and each generator's fakes resampled, $B=1{,}000$).}
    \label{fig:layers}
\end{figure}

\paragraph{The signal is strongest in intermediate layers.}
At the final layer, which semantic detectors read, the DINOv3 class token barely separates real from generated images (mean AUROC 0.52 \ci{0.50}{0.54}; \cref{fig:layers}a). At layers 10--16 it does (0.74--0.78; layer 16: 0.78 \ci{0.75}{0.80}), a gain of 0.26 \ci{0.24}{0.28} over the final layer on the same images. DINOv2 shows the same shape at a lower level: 0.68 at layer 14, and 0.38 at its final layer, which is \emph{below} chance. Chameleon follows the same curve (DINOv3 layer 16: 0.78 \ci{0.74}{0.81}; final layer 0.63; \cref{fig:layers}b). The full curves agree with the 470-image pilot from which \dv{}'s layers 8--18 were chosen: for DINOv3, which \dv{} uses, no layer outside that range is better on either benchmark (for DINOv2 on Chameleon, layer 22 is marginally better, 0.73 against 0.72 at layer 14). The drop at the end of the network is not specific to self-supervised training. CLIP, trained to match images with captions, has the same shape on the same images and copies: 0.85 at layer 14 \ci{0.84}{0.87} and 0.59 at its final layer, a drop of 0.26 \ci{0.24}{0.28}, as large as DINOv3's. Whatever their objective, the final layers of all three models discard most of the difference between an image and its copy, while intermediate layers keep it. We did not test a backbone trained with labels.

The signal is uneven across generators. At layer 16 it is strongest for Midjourney (0.99) and the three generators that decode through SD1's autoencoder (0.89 each), weaker for VQDM (0.80), GLIDE (0.68) and ADM (0.63), and absent for BigGAN (0.47), which has no autoencoder. That is the ordering the premise predicts.

\paragraph{Patch tokens and pixels mislead.}
Patch-level distances and PSNR reward images that are easy to reconstruct, whatever their origin. A blurred real image reconstructs well. When real images are blurred to the PSNR of their own copy, PSNR rates them as generated with AUROC 0.98 \ci{0.97}{0.99}, and so does the DINOv3 patch mean at layer 2 (0.94; \cref{fig:layers}c). The intermediate class token does the opposite. It rates the blurred images as \emph{less} generated than the originals (0.07--0.38 across layers 8--18 of both backbones; DINOv3 layer 16: 0.31 \ci{0.28}{0.34}). This is not neutral but inverted: blurring makes an image look more real to the score, so blurring a generated image could also hide it, a boundary we test in \cref{sec:wild}. At the same time the class token separates a simulated fake $\Rc(x)$ from $x$ almost perfectly (0.99). This resistance to blur is a property of DINO, not of intermediate layers in general. CLIP's intermediate class token separates GenImage's generated images better than DINOv3's, by 0.075 \ci{0.051}{0.098} at its best layer. But it also rates the blurred reals as generated: 0.83 \ci{0.81}{0.86} at layer 12 and 0.73 \ci{0.70}{0.76} at layer 14, reaching chance only from layer 16 on (0.53). Part of CLIP's stronger signal is therefore a smoothness signal, and on Chameleon CLIP is weaker than DINOv3 (best layer 0.72 against 0.78). This is why \dv{} reads DINOv3, why it reads the class token of intermediate layers, and why its training rows include the blur-matched counterfactual: a detector built on PSNR or early patch tokens would, in part, be a sharpness detector. Patch distances also invert on GenImage's 128-pixel BigGAN images (AUROC 0.04 at layer 16), which move much more under the copy than real photographs.

\paragraph{The copy should be made at native resolution.}
\Cref{fig:layers}d changes only how the copy is made. A $\times0.75$ downscale before the copy costs 0.03 AUROC on both backbones (DINOv3 $-0.032$ \ci{-0.051}{-0.016}). Squashing the image to $256\times256$, as SimLBR does before reading it, costs DINOv3 0.075 \ci{-0.095}{-0.055}. Most of this loss comes from the generator whose images are shrunk the most: Midjourney (1024 pixels, shrunk four-fold) falls from 0.99 to 0.73, while the 512-pixel SD generators lose at most 0.03. For DINOv2 the squash makes no measurable difference ($-0.005$ \ci{-0.025}{0.015}). Re-saving both the view and its copy as JPEG q95 \emph{raises} AUROC by 0.08 on both backbones, presumably because it removes fine detail in which every copy differs from its original. Resizing changes the image's pixel grid, so a strongly resized image is no longer the decoder's own output. This is why \dv{} copies a native-resolution crop.

\paragraph{Finding.} \textit{Without training, the fixed point is visible to frozen foundation models, self-supervised or trained on image--text pairs, mainly in their intermediate layers; a head trained on copies can still read a difference at the final layer (\cref{app:ablation}). In DINO these layers do not mistake smoothness for generation, while in CLIP they partly do. The signal survives mild downscaling but not strong downscaling.}

%% file: sections/decoder.tex
\section{Whose trace is it?}
\label{sec:decoder}

\textit{Question: is the d\'ej\`a vu a property of latent diffusion in general, or of the exact decoder that produced the image?}

If a generated image is a near fixed point of its generator's autoencoder, the trace should be strongest under that exact decoder and weaker under any other, even a near-identical one. We test this without any learned head. Each image and its matched real are copied with every autoencoder in a bank, and each copy is scored by the training-free signal of \cref{sec:signal}. The clean test is Synthbuster, where every generated image comes from a known generator, real and generated images are stored losslessly, and their sizes match image by image (\cref{sec:protocol}).

\paragraph{Generators and their decoders.}
Synthbuster's generators span every case our analysis distinguishes. Stable Diffusion~2 decodes with exactly the autoencoder \dv{} uses (ft-MSE). Stable Diffusion~1.3 and 1.4 share its encoder but decode with SD1's original decoder, from which ft-MSE was fine-tuned. SDXL \citep{podell2024sdxl} has its own autoencoder; our bank holds the version distributed as \texttt{stabilityai/sdxl-vae}. DALL$\cdot$E~3 decodes Stable Diffusion's latent space with its own diffusion decoder, distilled to two steps with consistency distillation \citep[App.~B]{betker2023dalle3,song2023consistency}; OpenAI has released this decoder. DALL$\cdot$E~2 \citep{ramesh2022dalle2} and GLIDE \citep{nichol2022glide} generate in pixel space and have no autoencoder. For Midjourney~v5 and Adobe Firefly the architecture is not public.

\paragraph{Predictions, stated before scoring.}
Copying each generator's images with its own decoder should give the strongest training-free signal: SD1's original decoder for SD~1.3 and 1.4, ft-MSE for SD~2, SDXL's autoencoder for SDXL, and the released consistency decoder for DALL$\cdot$E~3. In particular, DALL$\cdot$E~3 should show a strong signal under the consistency decoder and a weak one under ft-MSE and SD1's decoder, although all three share the same encoder. We cannot verify that the DALL$\cdot$E~3 service used to build Synthbuster runs exactly the released decoder.

\paragraph{The bank.}
The bank holds SD1's original decoder, ft-MSE, SDXL, Kandinsky's MoVQ, a FLUX-family autoencoder, DC-AE and the consistency decoder (\cref{app:impl}). For each generator we copy 200 generated images and their 200 matched test reals with every autoencoder, without the JPEG re-save, and score each copy by how little the DINOv3 class token moves (z-scored on the real images, layers 10--18). The threshold is set in-sample, at the 95th percentile of the generator's own 200 matched reals. Because real and generated images have the same sizes, resolution cannot produce a signal here.

\input{sections/tab_sbbank}

\paragraph{Three of the five predictions hold.}
With downscaled reals (\cref{tab:sbbank}, top), SD1's original decoder detects 99.5\% of SD~1.3 and 99.0\% of SD~1.4 images, while ft-MSE, which shares its encoder, detects 21--26\%. ft-MSE detects 95.0\% of SD~2 images, and no other autoencoder more than 41\%. When each image is attributed to the autoencoder under which it moves least, 100\%, 98.5\% and 94.5\% of these images go to their own decoder.

\paragraph{SDXL: two versions of one autoencoder.}
The SDXL prediction fails as specified: the SDXL autoencoder in our bank detects 7.0\% of SDXL images, and no autoencoder more than 12\%. We then looked for the cause, so what follows is post hoc. Synthbuster's generation parameters name samplers from Stability AI's own code base, which decodes with the autoencoder stored in the model checkpoint. The official SDXL~1.0 repository ships two versions of the autoencoder, in folders named \texttt{vae} and \texttt{vae\_1\_0}; ours is bit-identical to the first, and the weights of the two differ only slightly (\cref{app:impl}). With the second version, the fixed-point score detects 100\% of Synthbuster's SDXL images (AUROC 1.000) and attributes 99.5\% of them to it, while it detects at most 23.5\% of any other generator's images. Two versions of one autoencoder thus detect 7\% and 100\% of the same images; we add the second version to the bank, so that it holds eight autoencoders.

\paragraph{DALL$\cdot$E~3 and the other generators.}
The DALL$\cdot$E~3 prediction holds only nominally: the consistency decoder gives the highest TPR in its row (34.5\%, against 32.5\% for ft-MSE and FLUX), but by AUROC it is behind three other autoencoders. This decoder adds noise at each of its two steps, so even the exact decoder cannot return an image to itself; this may explain the weak signal, but we have not tested it. No autoencoder approaches the matched levels on the pixel-space generators (at most 44\%) or on Firefly (at most 8.5\%). Midjourney~v5 behaves like an ft-MSE image (91.0\%), as GenImage's Midjourney does; we do not know its architecture and draw no conclusion (\cref{app:sbmore}).

\paragraph{Under both matchings.}
With cropped reals (\cref{tab:sbbank}, bottom), the exact decoder remains the strongest autoencoder for all four Stable Diffusion versions: it detects 100\% of SD~1.3 and 1.4 images, 94.5\% of SD~2 and 100\% of SDXL. Native-resolution crops of camera photographs change more under every autoencoder than downscaled ones, so the other autoencoders detect more: the near-identical versions 16.5--42.0\%, and unrelated autoencoders up to 81.0\% (SDXL's on SD~2). We therefore measure specificity as the exact decoder's AUROC margin over the best other autoencoder, re-chosen in every bootstrap draw. The margin is 0.15--0.34 with downscaled reals and 0.03--0.24 with cropped reals, and its 95\% interval excludes zero for all four generators under both matchings; the smallest is SD~2 with cropped reals, 0.030 \ci{0.016}{0.049}. With cropped reals the DALL$\cdot$E~3 prediction fails outright (\cref{app:sbmore}).

\paragraph{On GenImage and on web images.}
GenImage cannot give as clean a test, because its real images are JPEG and its generated images PNG; copied as stored, the JPEG reals move more under every autoencoder, so we re-save every view as JPEG q95 before the copy (\cref{app:bankorig}). SD1's original decoder is then still the best single autoencoder for SD~1.4, SD~1.5 and Wukong (66\% of the 600 images pooled, against 52\% for ft-MSE), but the margin is small, and the re-save itself perturbs the decoder's output, so GenImage supports the account only weakly. No autoencoder in the bank matches Chameleon, whose real and generated images are both mostly JPEG at quality 75 (\cref{sec:chameleon}): the best single one (DC-AE) detects 37\%, the bank maximum 34\%, and the SDXL version that matches Synthbuster exactly 16\%.

\paragraph{Changing the decoder that makes the copies.}
If the trace belongs to the exact decoder, a \dv{} trained on copies from SD1's original decoder instead of ft-MSE should gain on SD1 images and lose where ft-MSE is the matched decoder. We wrote down four such predictions before training and retrained both variants with only the copy decoder changed (three seeds; \cref{tab:e8}). All four hold: on Synthbuster, SD~1.3 and 1.4 gain 5--8 points while SD~2, whose decoder is ft-MSE, loses about 2, and on GenImage, Midjourney, which ft-MSE matches best, loses 3--8 (\cref{app:e8}). The copy decoder also shifts whole benchmarks in opposite directions, so no single one is best everywhere; the rest of this paper uses ft-MSE, a choice made at the start of the study.

\paragraph{Finding.} \textit{Where it can be tested cleanly, the training-free d\'ej\`a vu is a property of the generator's exact decoder. Under both ways of matching the real photographs, the generator's own decoder separates the images of SD~1.3, 1.4, 2 and XL from real photographs better than any of seven other autoencoders (for SDXL, after identifying its version post hoc). It detects 94--100\% of them, while a near-identical version of the same autoencoder detects 7--42\%. No autoencoder we tried matches web images.}

%% file: sections/tab_sbbank.tex
\begin{table}[t]
\caption{\textbf{Decoder bank on Synthbuster} (training-free; 200 generated images and their 200 matched test reals per generator, no re-save). Top: reals downscaled to each generated image's size (pre-specified). Bottom: the same generated images against cropped reals (\cref{sec:protocol}), for the five generators with a pre-specified decoder. TPR (\%) at 5\% FPR of the fixed-point score (DINOv3 class token, layers 10--18) under each autoencoder; the threshold is the 95th percentile of the generator's own matched reals. Bold: the generator's own decoder as pre-specified. $^\dagger$Added post hoc, after the SDXL prediction failed; underlined: the SDXL version Synthbuster's images match. SD1: SD1's original decoder; MoVQ: Kandinsky~2.1; Cons.: OpenAI's consistency decoder. With 200 images per cell, a 95\% binomial interval spans at most $\pm7$ points.}
\label{tab:sbbank}
\centering\footnotesize\setlength{\tabcolsep}{4pt}
\begin{tabular}{@{}lrrrrrrrr@{}}
\toprule
Generator & SD1 & ft-MSE & SDXL 0.9 & SDXL 1.0$^\dagger$ & MoVQ & FLUX & DC-AE & Cons. \\
\midrule
\multicolumn{9}{@{}l}{\emph{Downscaled reals}} \\
SD~2 & 16.0 & \textbf{95.0} & 41.0 & 15.5 & 10.0 & 21.5 & 11.5 & 13.5 \\
SD~1.3 & \textbf{99.5} & 21.0 & 1.0 & 6.0 & 9.0 & 19.5 & 13.5 & 0.5 \\
SD~1.4 & \textbf{99.0} & 26.0 & 3.5 & 10.0 & 4.0 & 21.5 & 22.5 & 0.5 \\
DALL$\cdot$E 3 & 22.0 & 32.5 & 31.5 & 15.0 & 7.0 & 32.5 & 2.0 & \textbf{34.5} \\
SDXL & 1.5 & 12.0 & \textbf{7.0} & \underline{100.0} & 1.0 & 0.0 & 11.0 & 0.5 \\
\midrule
DALL$\cdot$E 2 & 16.0 & 22.0 & 3.0 & 14.0 & 18.5 & 3.5 & 8.0 & 44.0 \\
GLIDE & 10.5 & 10.0 & 1.5 & 9.5 & 2.0 & 23.5 & 41.5 & 0.0 \\
Midjourney v5 & 41.5 & 91.0 & 42.5 & 23.5 & 19.5 & 63.5 & 22.5 & 22.0 \\
Firefly & 4.0 & 8.5 & 2.0 & 3.0 & 0.0 & 8.0 & 0.5 & 5.5 \\
\midrule
\multicolumn{9}{@{}l}{\emph{Cropped reals}} \\
SD~2 & 42.0 & \textbf{94.5} & 81.0 & 34.0 & 30.5 & 3.5 & 2.5 & 17.5 \\
SD~1.3 & \textbf{100.0} & 16.5 & 44.5 & 29.0 & 68.0 & 0.0 & 3.0 & 14.0 \\
SD~1.4 & \textbf{100.0} & 16.5 & 40.5 & 26.5 & 60.5 & 0.0 & 1.5 & 16.5 \\
DALL$\cdot$E 3 & 41.0 & 24.5 & 78.5 & 35.0 & 11.5 & 2.5 & 0.5 & \textbf{38.0} \\
SDXL & 8.5 & 6.5 & \textbf{30.5} & \underline{100.0} & 4.5 & 0.0 & 3.5 & 1.0 \\
\bottomrule
\end{tabular}
\end{table}

%% file: sections/zerofake.tex
\section{Does a detector trained without fakes work?}
\label{sec:zerofake}

\textit{Question: can a detector that has never seen a generated image match detectors trained on thousands of them?}

\begin{table}[t]
\caption{\textbf{Main results.} TPR (\%) at a 5\% and 1\% false-positive rate set on real calibration images only, and false-positive rate (\%) on degraded real images at the 5\% threshold (lower is better). GenImage: mean and worst over eight generators, all images re-saved as JPEG q95. Synthbuster: mean and worst over nine generators, with reals downscaled to each generated image's size (\cref{sec:protocol}; cropped reals in \cref{tab:sbcrop}). ``Fakes'': generated images used in training. RIGID with normalised noise: noise added to the normalised input (\cref{app:rigid}). Learned methods: mean over 3 seeds. $^\dagger$AEROBLADE is scored on a fixed stratified subset (4{,}000 images per benchmark); every method scored on that subset is in \cref{app:subset}. $^\ddagger$The first version of \dvI{}, which reads the final-layer class token; it was replaced by layers 8--18 (\cref{sec:signal}) and is shown for reference only (\cref{sec:finallayer}). Bold: best single detector per column, excluding this row; the fused detector (last row) is discussed in \cref{sec:fusion}. Paired intervals for the key comparisons are in the text.}
\label{tab:main}
\centering
\footnotesize
\setlength{\tabcolsep}{2.8pt}
\begin{tabular}{@{}lrrrrrrrrrrr@{}}
\toprule
 & & \multicolumn{4}{c}{GenImage} & \multicolumn{3}{c}{Chameleon} & \multicolumn{3}{c}{Synthbuster} \\
\cmidrule(lr){3-6}\cmidrule(lr){7-9}\cmidrule(l){10-12}
Method & Fakes & TPR@5 & worst & TPR@1 & FPR$_\text{deg}$ & TPR@5 & TPR@1 & FPR$_\text{deg}$ & TPR@5 & worst & FPR$_\text{deg}$ \\
\midrule
\multicolumn{12}{@{}l}{\textit{Trained with generated images}} \\
SimLBR (released) & 162k & 85.3 & 45.0 & \textbf{74.2} & 4.2 & \textbf{84.0} & \textbf{75.6} & 7.7 & 80.3 & \textbf{41.9} & 4.5 \\
SimLBR (retrained) & 10k & 83.8 & 41.0 & 73.0 & 3.7 & 81.1 & 68.0 & 6.4 & 81.3 & 40.5 & 5.3 \\
Supervised probe & 10k & 80.3 & 32.4 & 67.9 & \textbf{3.6} & 73.2 & 55.4 & 4.9 & 78.4 & 31.6 & 5.2 \\
\midrule
\multicolumn{12}{@{}l}{\textit{Training-free}} \\
RIGID, DINOv2 (pixel noise) & 0 & 32.1 & 7.2 & 14.7 & 19.2 & 4.6 & 0.2 & 19.5 & 5.4 & 0.5 & 16.5 \\
RIGID, DINOv2 (norm.\ noise) & 0 & 38.4 & 11.2 & 19.6 & 21.6 & 2.9 & 0.3 & 28.4 & 4.5 & 0.0 & 13.4 \\
RIGID, DINOv3 & 0 & 36.6 & 18.8 & 18.8 & 17.7 & 29.5 & 10.4 & 17.3 & 26.6 & 5.6 & 17.1 \\
AEROBLADE$^\dagger$ & 0 & 56.8 & 10.0 & 25.1 & 48.0 & 5.2 & 0.8 & 22.9 & 8.9 & 0.0 & 25.0 \\
\dv-TF (ours) & 0 & 43.6 & 0.8 & 22.0 & 5.5 & 27.7 & 7.3 & \textbf{0.8} & 23.5 & 0.8 & 4.1 \\
\midrule
\multicolumn{12}{@{}l}{\textit{Trained on real images only}} \\
Reconstruction probe & 0 & 81.0 & 50.9 & 64.7 & 9.9 & 63.0 & 38.1 & 12.5 & 70.1 & 32.4 & 10.9 \\
\dvI{} (ours) & 0 & \textbf{87.5} & \textbf{64.2} & 73.1 & 5.0 & 33.9 & 14.8 & 2.0 & \textbf{84.5} & 35.1 & 3.0 \\
\dvP{} (ours) & 0 & 78.6 & 36.2 & 64.5 & 4.0 & 49.2 & 27.0 & 2.5 & 74.7 & 18.9 & \textbf{2.7} \\
\quad\textcolor{black!60}{\dvI{}, final layer$^\ddagger$} & \textcolor{black!60}{0} & \textcolor{black!60}{79.5} & \textcolor{black!60}{58.3} & \textcolor{black!60}{64.2} & \textcolor{black!60}{3.6} & \textcolor{black!60}{61.2} & \textcolor{black!60}{37.0} & \textcolor{black!60}{3.9} & \textcolor{black!60}{85.7} & \textcolor{black!60}{69.7} & \textcolor{black!60}{4.0} \\
\midrule
\dvP{} + SimLBR, fused & 162k & 91.5 & 58.2 & 82.5 & 4.1 & 87.1 & 74.3 & 3.5 & 91.9 & 62.8 & 3.0 \\
\bottomrule
\end{tabular}
\end{table}

\paragraph{On GenImage, zero fakes are enough.}
We planned \dvP{} as the method and trained \dvI{} from the start as its control, to test whether reading the copy helps. Both are reported everywhere. The text leads with \dvI{} because it is the stronger zero-fake detector on GenImage. This choice was made after the GenImage results, but both variants were frozen before Synthbuster was scored, so Synthbuster (\cref{sec:synthbuster}) tests the choice on data it never saw; there, \dvI{}'s lead over SimLBR depends on how the real photographs are matched. \dvI{} detects 87.5\% of GenImage fakes at a 5\% false-positive rate (\cref{tab:main}). This is higher than the released SimLBR head, which was trained on 162k generated images (+2.3 points \ci{0.9}{3.7}), and much higher at the worst generator (64.2 vs.\ 45.0). It is above SimLBR retrained on 10k generated images (83.8), and above SimLBR trained on 1.25k generated images (72.1; \cref{app:scale}). It needs no copy at test time, so it costs the same single DINOv3 pass as SimLBR (58\,ms per image on our hardware; \cref{app:impl}). Against the AlignedForensics-style reconstruction probe, which also uses no generated images, \dvI{} gains 6.5 points \ci{5.1}{8.1} on the mean and 13.3 \ci{9.6}{16.9} at the worst generator, a difference in three factors at once: intermediate layers, native-resolution copies and counterfactual reals. Changing one factor at a time (\cref{app:ablation}), the intermediate layers add 8.0 points on GenImage \ci{6.5}{9.4} over the same head on the final layer, though not on Synthbuster, and the counterfactual rows mainly control false alarms. The same choices hurt on Chameleon, where the reconstruction probe, which reads squashed final-layer features, is 28.9 points \ci{26.3}{31.5} ahead. Most of that gap is the layer: read at the final layer, \dvI{} detects 61.2\% of Chameleon's generated images, close to the reconstruction probe's 63.0\% (\cref{tab:main}, grey row; \cref{sec:finallayer}). \Cref{sec:wild} examines why the intermediate-layer trace does not survive web processing.

\paragraph{Both training-free baselines fall far behind.}
\dvI{} exceeds RIGID by 55.4 points \ci{53.4}{57.5} (48.9 \ci{46.9}{50.9} with the stronger noise convention) and AEROBLADE, on its subset, by 28.6 \ci{23.1}{35.0}. RIGID is near chance on latent diffusion images (6--8\%), with or without the JPEG re-save (\cref{app:rigid}). AEROBLADE reaches 71--84\% on most generators but flags 48\% of degraded real photographs, because a slightly blurred photograph reconstructs well; \dv-TF keeps that rate at 5.5\%.

\paragraph{Counterfactual reals prevent false alarms.}
Removing the blur-matched counterfactual rows raises \dvP{}'s TPR on GenImage by 8.6 points \ci{8.0}{9.2}, but it buys that gain by flagging soft images. Without the counterfactual rows, \dvP{}'s false alarms on degraded real photographs rise from 4.0\% to 39.9\% on GenImage and from 2.5\% to 38.9\% on Chameleon. The fused detector inherits the problem (20.6\% and 25.8\%, against 4.1\% and 3.5\%). For \dvI{} the effect is smaller but has the same sign: from 5.0\% to 29.8\% on GenImage, 2.0\% to 12.8\% on Chameleon and 3.0\% to 16.8\% on Synthbuster (\cref{app:ablation}).

\paragraph{The image alone beats the pair, except on Chameleon.}
Reading the copy alongside the image does not help on GenImage: \dvP{} is 9.0 points \ci{$-10.1$}{$-7.9$} below \dvI. On Chameleon the order reverses, and \dvP{} is 15.1 points \ci{12.9}{17.2} above \dvI. On Synthbuster \dvP{} is again behind, by 9.8 points (5.1 with cropped reals; \cref{app:sbmore}). \dvP{} reads how the image moved, and \dvI{} only where it is. One plausible reading, which we have not tested, is that movement transfers better to the unknown generators and processing of web images, while position alone suffices on curated benchmarks. On Chameleon both variants are far below every detector trained with generated images.

\paragraph{Where it works, the autoencoder matters.}
\dvI{}'s gains are largest on generators that decode through a Stable Diffusion autoencoder (99.1--99.5\% on SD~1.4, SD~1.5 and Wukong). A matched decoder also helps the training-free scores: AEROBLADE detects 90\% of SD~1.4 images with SD1's original autoencoder and 33\% with the fine-tuned one, and on Synthbuster the exact decoder separates the classes almost perfectly (\cref{sec:decoder}). GenImage's ADM and BigGAN images are also detected (67\% and 84\%), but the fixed point does not explain this. BigGAN has no autoencoder and its images do not move less than real ones under the copy (AUROC 0.47 at layer 16; \cref{sec:where}), and ADM's training-free signal is weak (0.63). These detections come from other differences that the trained head picks up, and should not be read as evidence for the premise. On GenImage, resolution is one candidate; on Synthbuster, where resolution is matched, \dvI{} still detects the pixel-space DALL$\cdot$E~2 (\cref{sec:synthbuster}).

\paragraph{Finding.} \textit{When the image is unprocessed, a detector trained on real photographs and their copies matches a detector trained on 162k generated images on GenImage. On modern generators it is competitive, ahead or behind depending on how the real photographs are matched (\cref{sec:synthbuster}), and on web images it is far behind.}

%% file: sections/synthbuster.tex
\section{Does the comparison hold without the size shortcut?}
\label{sec:synthbuster}

\textit{Question: on recent generators, evaluated so that size cannot separate the classes, how does a detector trained without generated images compare?}

GenImage's generators are dated, and Chameleon cannot be evaluated without first removing a size shortcut (\cref{sec:chameleon}). Synthbuster pairs every generated image with a real photograph of the same size, in two ways: downscaled, as pre-specified, and cropped at native resolution, added later (\cref{sec:protocol}). Detectors and heads are those of \cref{sec:zerofake}, frozen before any Synthbuster image was scored; nothing is retrained. Every interval resamples RAISE photographs, because each contributes nine matched reals and up to nine generated images.

\begin{figure}[t]
\centering
\includegraphics[width=\linewidth]{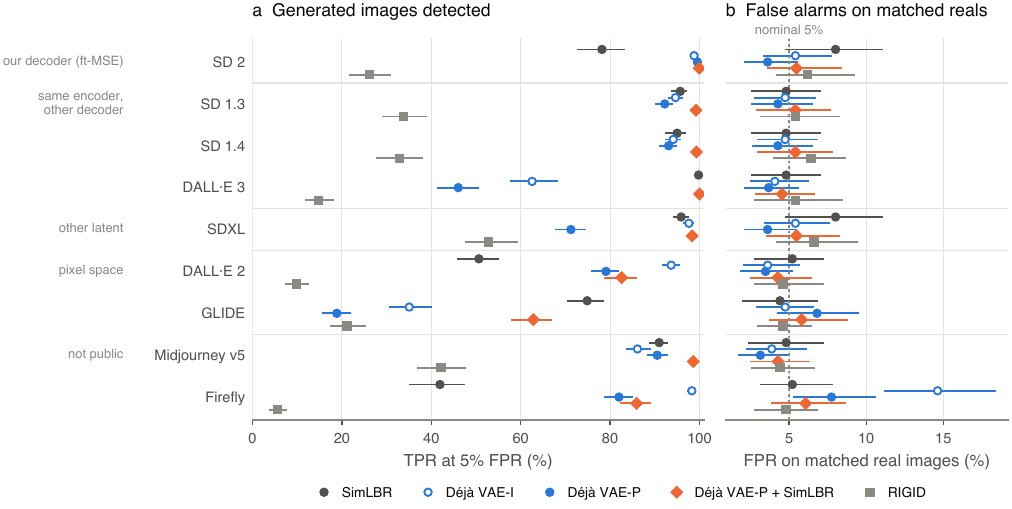}
\caption{\textbf{Synthbuster with downscaled reals.} (a) TPR at a 5\% false-positive rate per generator, with one threshold per method and seed set on the pooled calibration reals. Rows are grouped by how each generator decodes; ``our decoder'' is the ft-MSE autoencoder that makes \dv{}'s copies. (b) False-positive rate at the same threshold on each generator's matched test reals (nominal 5\%). SD~2 and SDXL images have the same sizes image by image, so their matched reals are identical. Points: mean over three seeds; bars: 95\% bootstrap intervals (calibration reals resampled by photograph; fakes and matched test reals resampled; $B=1{,}000$). The same comparison with cropped reals is in \cref{tab:sbcrop,app:sbmore}.}
\label{fig:synthbuster}
\end{figure}

\input{sections/tab_sbcrop}

\paragraph{The matching decides who leads.}
With downscaled reals, \dvI{} detects 84.5\% of Synthbuster images at a 5\% false-positive rate (\cref{tab:main}): 4.4 points above the released SimLBR head \ci{1.9}{7.1}, 6.1 above the supervised probe \ci{3.7}{8.5} and 14.6 above the reconstruction probe \ci{11.4}{17.9}. The lead does not survive the second matching. With cropped reals \dvI{} detects 68.2\%, 9.7 points below SimLBR \ci{$-13.2$}{$-6.1$} and level with the supervised probe ($-1.3$ \ci{$-5.1$}{2.8}), while SimLBR itself changes by 2.4 points (\cref{tab:sbcrop}). We therefore do not claim that \dvI{} is better than SimLBR on Synthbuster. The two matchings differ in the pixel scale of the real photographs, and \dvI{} is sensitive to it: on downscaled reals, the less a photograph was downscaled, the more generated it looks to \dvI{} (Spearman $\rho=-0.12$ over the 4{,}491 test reals; SimLBR $-0.04$).

\paragraph{Per generator.}
The per-generator pattern holds under both matchings (\cref{fig:synthbuster}a; intervals in \cref{app:sbmore}). \dvI{} is far ahead of SimLBR on DALL$\cdot$E~2 (+43.0 with downscaled and +30.9 with cropped reals), Firefly (+56.4 and +54.4) and SD~2 (+20.6 and +20.0), and far behind on DALL$\cdot$E~3 ($-37.2$ and $-73.5$) and GLIDE ($-39.8$ and $-58.2$). The matching decides SD~1.3, SD~1.4, SDXL and Midjourney, where the two detectors are within 5 points with downscaled reals and \dvI{} is 8--30 points behind with cropped reals. Two results constrain how the detector should be read. DALL$\cdot$E~2 has no autoencoder, yet \dvI{} detects 93.7\% of its images (78.3\% with cropped reals): the detector learns how real photographs respond to the copy and flags images that respond differently, whatever made them. And DALL$\cdot$E~3, which shares SD~2's latent space but has its own decoder, is detected at only 62.6\% (26.1\%): sharing the encoder is not enough.

\paragraph{False alarms on matched reals.}
One pooled threshold transfers across generator geometries. With downscaled reals, every detector's false-positive rate lies between 3.1\% and 8.0\% on each generator's matched test reals (\cref{fig:synthbuster}b), except \dvI{} on the reals matched to Firefly, the least downscaled photographs (14.6\%). The Firefly gain does not depend on this: with a threshold set on those reals, \dvI{} detects 93.7\% of Firefly images and SimLBR 40.4\%. With cropped reals, \dv{} sees the same view for all nine reals of a photograph, so its false-positive rate is the same for every generator (\dvI{} 7.3\%, \dvP{} 5.0\%; SimLBR 3.6--9.0\%), and rests on 499 distinct test photographs. On degraded real images \dvI{} flags 3.0\% and SimLBR 4.5\%.

\subsection{A head on the final layer}
\label{sec:finallayer}
The matching sensitivity belongs to the intermediate layers. The study's first version of \dvI{}, identical except that its head reads the final-layer class token (\cref{sec:signal}), was scored on Synthbuster post hoc (grey rows in \cref{tab:main,tab:sbcrop}). It detects 85.7\% with downscaled and 86.3\% with cropped reals, against 84.5\% and 68.2\% for layers 8--18; a linear head shows the same contrast (final layer 88.8--90.4\% under both matchings; layers 8--18 falling from 85.6\% to 52.4\%; \cref{app:ablation}). It flags slightly more matched test reals than SimLBR (7.4\% and 6.4\%, against 5.6\% and 5.3\%). At equal false-positive rates, with every threshold set on the matched test reals, it is still ahead, by 3.6 points \ci{0.3}{6.4} with downscaled and 7.2 \ci{4.7}{10.2} with cropped reals (AUROC $+0.013$ and $+0.036$, both intervals above zero).

It is not a better detector overall. It is behind SimLBR on GenImage (79.5\%) and Chameleon (61.2\%, though far ahead of \dvI{}'s 33.9\%), and less tied to the copy decoder (SD~2: 86.7\% instead of 98.8\%). Measured from each head's clean score under the processing of \cref{sec:wild}, it loses less to resizing and blur (a mild blur: $-15.5$ points against $-26.8$; a downscale with JPEG q75: $-14.5$ against $-22.5$), and its false alarms on processed reals stay at 2.2--8.0\% (6.6--18.6\%). A JPEG q75 re-save alone, however, costs it 15.0 points (9.7). It reads a more generic difference, less sensitive to how real photographs are resampled but not to compression, which we have not identified. Because it is post hoc on Synthbuster, we do not promote it to the main comparison.

\paragraph{Finding.} \textit{With size removed as a cue, how the real photographs are matched to the generated images decides whether a detector trained on real photographs and their copies leads or trails one trained on 162k generated images ($+4.4$ or $-9.7$ points). The sensitivity belongs to the intermediate layers: read at the final layer (post hoc), the same detector is ahead under both matchings. Under both, it is far ahead on DALL$\cdot$E~2, Firefly and SD~2 and far behind on GLIDE and DALL$\cdot$E~3.}

%% file: sections/tab_sbcrop.tex
\begin{table}[t]
\caption{\textbf{Synthbuster under two ways of matching the real photographs.} \emph{Downscaled} (pre-specified): each RAISE photograph is cropped to the aspect ratio of its paired generated image and resized to its exact size. \emph{Cropped} (added after the downscaled results): a native-resolution centre crop at the generated image's exact size, with no resampling. Generated images, calibration/test split and detectors are identical. TPR (\%) at 5\% FPR, mean and worst over the nine generators, and false-positive rate (\%) on the matched test reals at the same threshold; mean over 3 seeds. $^\ddagger$The first version of \dvI{}, reading the final-layer class token, scored on Synthbuster post hoc (\cref{app:ablation}).}
\label{tab:sbcrop}
\centering\footnotesize\setlength{\tabcolsep}{5pt}
\begin{tabular}{@{}lrrrrrr@{}}
\toprule
 & \multicolumn{3}{c}{Downscaled reals} & \multicolumn{3}{c}{Cropped reals} \\
\cmidrule(lr){2-4}\cmidrule(l){5-7}
Method & Mean & Worst & FPR & Mean & Worst & FPR \\
\midrule
SimLBR (released) & 80.3 & 41.9 & 5.6 & 77.9 & 36.1 & 5.3 \\
Supervised probe & 78.4 & 31.6 & 5.5 & 69.5 & 21.3 & 3.8 \\
Reconstruction probe & 70.1 & 32.4 & 5.5 & 52.8 & 17.3 & 4.9 \\
\dvI{} & 84.5 & 35.1 & 5.7 & 68.2 & 13.1 & 7.3 \\
\dvP{} & 74.7 & 18.9 & 4.5 & 62.7 & 8.7 & 5.0 \\
\quad\textcolor{black!60}{\dvI{}, final layer$^\ddagger$} & \textcolor{black!60}{85.7} & \textcolor{black!60}{69.7} & \textcolor{black!60}{7.4} & \textcolor{black!60}{86.3} & \textcolor{black!60}{71.1} & \textcolor{black!60}{6.4} \\
\midrule
\dvP{} + SimLBR, fused & 91.9 & 62.8 & 5.2 & 85.8 & 49.6 & 4.4 \\
\dvI{} + SimLBR, fused & 95.0 & 72.1 & 5.6 & 88.0 & 57.8 & 4.8 \\
\bottomrule
\end{tabular}
\end{table}

%% file: sections/wild.tex
\section{What breaks the d\'ej\`a vu?}
\label{sec:wild}

\textit{Question: why does a detector that matches SimLBR on GenImage fall to 34--49\% on Chameleon?}

Two explanations fit the premise. Web images may come from generators whose autoencoders \dv{} does not cover; no autoencoder in our bank matches Chameleon (\cref{sec:decoder}). And images on the web are resized and re-compressed after generation, which moves them off the autoencoder's fixed point. Here we test the second.

\paragraph{Processing after generation.}
We take 150 fakes each from SD~1.4, Wukong and Midjourney, plus 300 real GenImage images, and apply one post-processing step to every image before the standard pipeline: a resize by $\times0.5$, $\times0.75$ or $\times1.5$, a JPEG re-save at quality 75, or a $\times0.75$ resize followed by JPEG q75. We recompute each detector's threshold on reals processed the same way, and we also report the false-positive rate on processed reals if the threshold is left where it was.

\begin{figure}[t]
    \centering
    \includegraphics[width=0.86\linewidth]{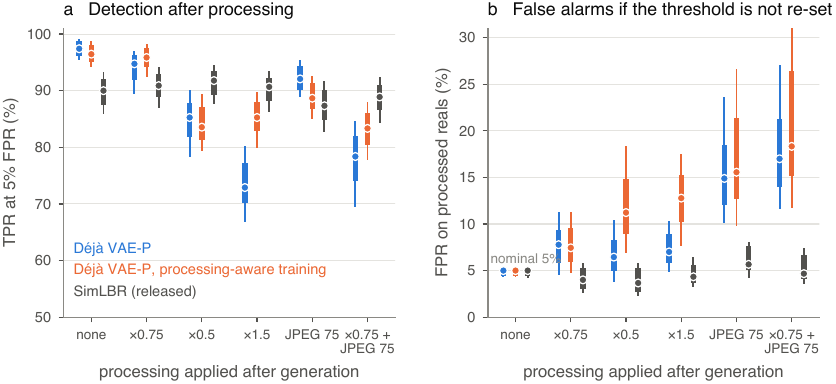}
    \caption{\textbf{One resize or re-compression moves generated images off the fixed point; a semantic detector does not notice.} \textbf{(a)} TPR@5\% after each post-processing step, with the threshold re-set on real images processed the same way. \textbf{(b)} False-positive rate on processed real images when the threshold set on unprocessed reals is kept; the dashed line is the nominal 5\%. Dots: mean over SD~1.4, Wukong and Midjourney (150 fakes each; 300 reals) and 3 seeds; thick and thin bars: 66\% and 95\% bootstrap intervals (reals and fakes resampled, $B=1{,}000$).}
    \label{fig:processing}
\end{figure}

On clean images \dvP{} detects 97.4\% of these fakes, against 90.0\% for SimLBR (\cref{fig:processing}). Mild processing barely matters: a $\times0.75$ resize costs \dvP{} 2.7 points. Stronger processing does. A $\times1.5$ upscale costs 24.5 points, and a $\times0.75$ resize followed by JPEG q75 costs 19.0 (\dvI: 22.5). SimLBR stays between 87.3\% and 91.8\% under every condition. Processing also shifts the score distribution of real images. If the threshold is left at its unprocessed value, JPEG q75 real photographs are flagged at 14.9--18.6\% instead of 5\%, while SimLBR stays at 3.7--5.7\%.

Chameleon shows the same pattern in the wild. Its fakes whose sides are not multiples of 8 pixels have almost certainly been resized after generation. \dvP{} detects 42\% of these, against 75\% of the fakes whose sides are both multiples of 64, which look like untouched generator output. SimLBR barely changes between the two groups (83\% vs.\ 87\%).

A mild blur is enough. Blurring generated images with the same Gaussian blur as our degraded real images ($\sigma\sim\mathcal U(0.5,1.5)$), with the threshold left where the calibration reals put it, lowers \dvI{} from 87.9\% to 61.1\% and \dvP{} from 78.1\% to 55.6\% on 1{,}600 GenImage fakes, but SimLBR only from 85.8\% to 77.7\%. This is the other side of the blur inversion of \cref{sec:where}: the score that refuses to call a blurred real image generated also lets a blurred generated image pass.

\paragraph{Processing-aware training does not fix it.}
Retraining \dv{} with every training row randomly resized and re-compressed recovers part of the loss on the controlled grid ($\times1.5$: 72.9\% $\to$ 85.3\%), but not on Chameleon (48.1\% vs.\ 49.2\%), and it makes the detector more sensitive to processing itself: at a fixed threshold, false alarms on resized real images roughly double (\cref{app:procaware}).

\paragraph{The layer and the decoder.}
Part of \dvI{}'s Chameleon gap is the choice of layers rather than processing: read at the final layer, it detects 61.2\% of Chameleon's generated images instead of 33.9\%, and resized and untouched ones alike, but that head is not robust to compression either (\cref{sec:finallayer}). An unmatched decoder does not explain the gap away: no autoencoder in our bank restores the training-free signal on Chameleon (\cref{sec:decoder}).

\paragraph{Finding.} \textit{The d\'ej\`a vu is a property of the generator's raw output. One resize with re-compression, or a mild blur, costs the trained detectors 19--27 points, while a semantic detector loses at most 8, and the processing-aware training we tried restores the trace only for the processing it was trained on.}

%% file: sections/chameleon.tex
\section{Is the in-the-wild benchmark measuring what we think?}
\label{sec:chameleon}

\textit{Question: how much of a detector's Chameleon score reflects generation, and how much reflects how the benchmark's files were made?}

\begin{figure}[t]
    \centering
    \includegraphics[width=0.86\linewidth]{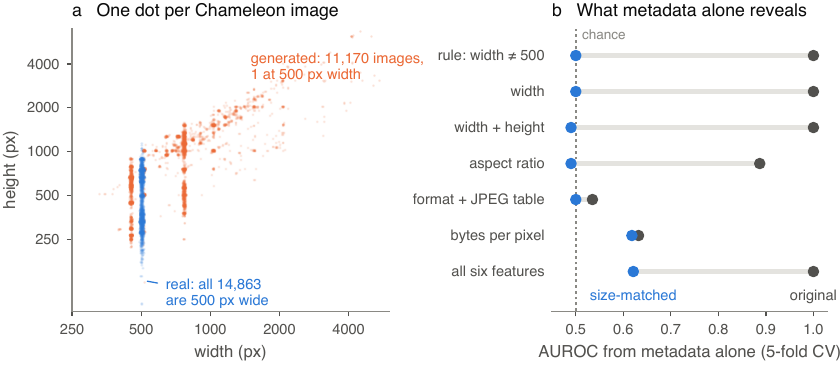}
    \caption{\textbf{Chameleon can be solved from image width alone.} \textbf{(a)} Width and height of every Chameleon image (one dot per image; log scales, slight jitter). All 14{,}863 real images are 500 pixels wide. \textbf{(b)} AUROC of gradient-boosted classifiers that see only file metadata (5-fold cross-validation), on the original benchmark (grey) and on our size-matched version (blue). The dashed line is chance. Size matching removes every geometric cue; the remaining signal (bytes per pixel, 0.62) reflects image content and is kept.}
    \label{fig:audit}
\end{figure}

\paragraph{A metadata shortcut.}
While grouping Chameleon images by size for \cref{sec:wild}, we found that every one of its 14{,}863 real images is exactly 500 pixels wide, while only 1 of its 11{,}170 generated images is (\cref{fig:audit}a). The rule ``width $\neq$ 500 $\Rightarrow$ generated'' detects 99.99\% of the fakes with no false positives, without reading a single pixel. The released real images also do not match the benchmark's description. \citet{yan2025sanity} describe Chameleon as ranging ``from 720P to 4K'' and state that images smaller than $448\times448$ were discarded. Yet no released real image has a short side of 720 pixels or more, and 57.3\% have a short side below 448 pixels; the corresponding figures for the generated images are 48.6\% and 6.7\%. This is consistent with the real images having been saved as fixed-width previews rather than as originals, but we cannot verify how the files were produced. We shared these observations with the authors before releasing this work. The real images also share every other file property we checked: all are baseline JPEG at quality 75 with 4:2:0 chroma subsampling and no EXIF data. Most fakes are JPEG at quality 75 too, so compression is \emph{not} the giveaway (format and JPEG table alone: AUROC 0.535). Geometry is: width, height and aspect ratio each separate the classes (AUROC 1.000, 1.000 and 0.887; real images are mostly landscape, with median height/width 0.72 against 1.5 for the fakes).

\paragraph{A size-matched Chameleon.}
To measure how much detectors depend on this, we build a version of Chameleon in which size carries no information. Each fake is given the geometry of a randomly drawn real image. It is centre-cropped to that image's aspect ratio, resized to exactly $500\times H$ pixels, and saved as baseline JPEG at quality 75 with 4:2:0 subsampling and no metadata. Real images are untouched. Metadata classifiers fall to chance on every geometric feature (\cref{fig:audit}b). The only remaining cue is bytes per pixel (AUROC 0.62), which depends on image content and which we keep.

\paragraph{Separating the shortcut from the processing.}
Size matching changes two things at once: it removes the shortcut, and it adds a resize and a JPEG re-save to every fake. Our own results in \cref{sec:wild} show that processing alone hurts detectors. We therefore add a \emph{processing-only control}: each fake receives the same resize factor and the same JPEG re-save as in the size-matched set, but keeps its own aspect ratio and full content. The three versions share every real image and every fake identity, so per-detector differences are paired. Original minus control is the cost of processing; control minus size-matched is the cost of losing the shortcut. The size-matched fakes are also centre-cropped (66\% of the image area kept, median), so the second difference also includes the loss of content and the change of aspect ratio.

\paragraph{Processing costs 5--11 points; the shortcut costs 0--4.}
Every detector loses ground on size-matched Chameleon (SimLBR 84.0\% $\to$ 75.8\%, \dvP{} 49.2\% $\to$ 37.6\%, RIGID-DINOv3 29.5\% $\to$ 15.0\%). The decomposition in \cref{fig:decomp} shows where the loss comes from. One extra resize and JPEG re-save costs SimLBR 5.8 points \ci{5.3}{6.3}, the supervised probe 6.5, \dvP{} 11.0 and RIGID 10.9. Removing the shortcut costs SimLBR 2.4 points \ci{1.9}{3.0} and RIGID 3.6, and costs the supervised probe and both \dv{} variants less than one point. For the reconstruction probe it is negative ($-2.6$): the reconstruction probe does slightly better without the shortcut.
Enlarged fakes do not explain these results: SimLBR detects fewer of the 38\% of size-matched fakes that had to be upscaled than of the downscaled ones (72\% vs.\ 78\%).

\paragraph{What this means for the benchmark.}
The detectors we test resize or crop every input to a fixed square, so the shortcut reaches them only through what that does to the image, which is why they depend on it so little. A detector that reads file metadata or aspect ratio, or one trained on data collected the same way, could exploit it fully. We recommend reporting Chameleon results on the size-matched version together with the processing-only control, and we release both as lists of transformations applied to the original files.

\paragraph{Finding.} \textit{Chameleon can be solved from image width, but resize-first detectors barely use that shortcut. What the benchmark really tests is robustness to one more round of web processing, which costs current detectors 5--11 points.}

%% file: sections/fusion.tex
\section{Is the d\'ej\`a vu still useful?}
\label{sec:fusion}

\textit{Question: when the fixed point is too fragile to stand alone, does it add anything to a semantic detector?}

\paragraph{Fusion improves the stronger detector on every benchmark.}
Adding \dvP{} to the released SimLBR head, by summing their scores after standardising each on real calibration images, raises TPR@5\% by 6.3 points on GenImage \ci{5.6}{7.1}, by 3.1 on Chameleon \ci{2.1}{4.1}, by 3.0 on size-matched Chameleon \ci{1.9}{4.3}, and by 11.7 on Synthbuster \ci{9.9}{13.7} (8.1 \ci{6.5}{9.6} with cropped reals; \cref{tab:main,fig:fusion,sec:synthbuster}). At the worst generator the gain is 13.3 points on GenImage \ci{11.1}{15.8} and 21.4 on Synthbuster \ci{15.5}{27.4}. False alarms on degraded real images do not rise: 4.1\% against 4.2\% on GenImage, and 3.5\% against 7.7\% on Chameleon.

\paragraph{The gain is complementarity, not ensembling.}
A gain from fusion could come from averaging any two detectors. We therefore compare with four control fusions, each pairing SimLBR with a different second detector. Fusing SimLBR with the supervised probe, a second semantic detector, gains much less: our fusion beats it by 6.3 points \ci{5.6}{6.9} on GenImage and 3.9 \ci{2.9}{4.9} on Chameleon. Our fusion also beats the RIGID fusion (+5.9 \ci{4.6}{7.0} and +11.4 \ci{9.5}{13.2}) and the reconstruction-probe fusion on Chameleon (+4.8 \ci{3.9}{5.8}). On GenImage it is only slightly better than the reconstruction-probe fusion on the mean (+0.8 \ci{0.1}{1.5}) and clearly worse at the worst generator ($-14.0$ \ci{$-16.4$}{$-11.6$}). It also beats the AEROBLADE fusion on the AEROBLADE subset (\cref{app:subset}). Synthbuster is where fusion helps most. There our fusion beats the supervised-probe and reconstruction-probe fusions by 8.5 \ci{7.2}{10.3} and 9.5 \ci{8.1}{11.2} points with downscaled reals and by 9.0 \ci{7.3}{10.8} and 9.7 \ci{8.1}{11.3} with cropped reals, and the RIGID fusion by 19.6 \ci{16.5}{23.2}, while fusing SimLBR with the supervised probe adds 3.1 points \ci{1.6}{5.0} with downscaled reals and nothing with cropped reals ($-1.0$ \ci{$-2.1$}{0.1}). The gain comes from DALL$\cdot$E~2, Firefly and SD~2, where SimLBR is weakest; on GLIDE the fusion falls to 62.8\% from SimLBR's 74.9\%.

\paragraph{Which variant to fuse.}
The two \dv{} variants trade places here as they do alone. On GenImage the \dvI{} fusion is better (by 2.7 points \ci{2.2}{3.1}; on Synthbuster by 3.1 \ci{2.3}{3.9}), and on Chameleon the \dvP{} fusion is better (by 4.8 \ci{4.0}{5.7}). \dvP{} was the planned method from the start of this study, so we report its fusion throughout rather than picking the better variant per benchmark. The counterfactual rows remain essential: fusing the variant trained without them raises GenImage TPR to 93.6\% but flags 20.6\% of degraded real images (25.8\% on Chameleon).

\paragraph{Finding.} \textit{The fixed point carries information that a semantic detector lacks. Fusing the two improves detection on all three benchmarks, including after the size shortcut is removed, and keeps false alarms on degraded real photographs at the base rate.}

%% file: sections/limitations.tex
\section{Limitations}
\label{sec:limitations}

\paragraph{What we do not claim.}
We do not claim a state-of-the-art in-the-wild detector: \dv{} alone is far below SimLBR on Chameleon. We do not claim that comparing an image with its copy is better than reading the image alone; the answer differs between benchmarks. We do not claim that training without generated images is new (\cref{sec:related}), and we make no claim about generators whose autoencoders are not in our bank.

\paragraph{Prior knowledge.}
\dv{} uses no generated images, but it uses the Stable Diffusion autoencoder, which is prior knowledge about how the test generators work; its GenImage advantage is largest on generators that decode through it. \dv{} trades generated training images for knowledge of the decoder, and no single copy decoder is best on every benchmark (\cref{sec:decoder}).

\paragraph{Layer selection.}
The layers were chosen from a training-free pilot on 2\% of the GenImage and 1.3\% of the Chameleon test fakes, after the final-layer version had been scored on both test sets (\cref{sec:signal}). The full layer curves do not change the choice (\cref{fig:layers}), and removing all 820 pilot images from the evaluation changes the TPR@5\% of SimLBR, \dvI{} and \dvP{} by at most 0.2 points. For the trained head, however, the choice does not transfer: on Synthbuster, the one benchmark held out from it, the final layer is as good with downscaled reals and far better with cropped reals (\cref{sec:finallayer}). A layer choice that is best everywhere remains open.

\paragraph{Benchmarks and baselines.}
GenImage is dated, and its fakes cluster at a few sizes \citep{grommelt2024fakejpeg}; we control its file-format bias but not its size bias. Chameleon is a single web-collected set. Synthbuster has 1{,}000 images per generator and real images from one camera dataset, and matching them to the generated images' size has no neutral solution (\cref{sec:protocol,sec:synthbuster}); a benchmark whose real images are produced at the generated images' sizes would avoid the choice. AEROBLADE is evaluated on a fixed subset and a 256-pixel crop to keep its cost comparable, and RIGID's noise convention is unspecified, so we report both. We do not compare with detectors that bring their own backbone and training data, such as AIDE \citep{yan2025sanity}, DRCT \citep{chen2024drct}, DDA \citep{chen2025dda} or SPAI \citep{karageorgiou2025spai}: our comparisons hold backbone, head and protocol fixed so that differences can be attributed to the signal, and our numbers are not a ranking of published detectors.

\paragraph{Decoder identification.}
The bank results depend on knowing a generator's autoencoder. For SDXL we identified the matching version only after the pre-specified test failed; we cannot verify that the DALL$\cdot$E~3 service ran the released consistency decoder; and Midjourney's and Firefly's architectures are not public. The bank's thresholds are set in-sample, and unrelated autoencoders can detect up to 81\% of a generator's images, so a high rate points to a decoder only when it is the highest in its row.

\paragraph{Compute.}
All experiments ran on one laptop (Apple M5, 16\,GB unified memory, PyTorch MPS backend). At test time, \dvI{} costs one DINOv3 pass (58\,ms per image), the same as SimLBR, and \dvP{} and \dv-TF add a copy and a second pass (329\,ms; \cref{app:impl}). Scoring a full benchmark, which also copies a blurred version of every test real, takes 2.5--4.5 hours, and the heads train on the CPU in minutes. Memory limited the training set (5{,}000 real photographs for \dv{}, 10{,}000 images per class for the retrained baselines); the retrained SimLBR gains at most 1.7 points from 5{,}000 to 10{,}000 per class (\cref{app:scale}). We did not vary the training-set size of \dv{}.

\section{Conclusion}
\label{sec:conclusion}

A generated image's d\'ej\`a vu is real but narrow. Without training, a frozen foundation model sees it mainly in intermediate layers and much less at the final layer that detectors usually read; a head trained on autoencoder copies finds a difference there too, but one that is no more robust to compression. Where it can be tested cleanly, it belongs to the generator's exact decoder: on Synthbuster, exact decoders catch 94--100\% of the images of four Stable Diffusion versions without any training, and near-identical versions of the same decoders 7--42\%. A detector trained on real photographs and their copies matches a detector trained on 162k generated images on GenImage and complements it in fusion. On a size-controlled modern benchmark it leads or trails depending on how the real photographs are matched; read at the final layer (post hoc), it leads under both. One resize with re-compression, or a mild blur, costs the trained detectors 19--27 points, an unmatched decoder removes most of the training-free signal, and on web images no autoencoder we tried restores it. Two benchmark lessons are independent of the trace. A widely used in-the-wild benchmark can be solved from image width, although resize-first detectors barely use that shortcut. And when real photographs must be matched to the size of generated images, the matching itself can reverse a detector comparison, so benchmarks built this way should report more than one matching. We hope the protocol, the size-matched Chameleon and the two matched versions of Synthbuster help the next detector be measured against what the open web actually does to images.

%% file: sections/appendix.tex
\section{Implementation details}
\label{app:impl}

\paragraph{Backbones.} DINOv3 ViT-L/16 (\texttt{facebook/dinov3-vitl16-pretrain-lvd1689m}) with SimLBR's normalisation (mean $(0.430, 0.411, 0.296)$, std $(0.213, 0.156, 0.143)$) at $256\times256$ input; hidden state $\ell$ is the output of block $\ell$, and the class token is token 0. DINOv2 ViT-L/14 (\texttt{facebook/dinov2-large}) with ImageNet normalisation at $224\times224$. CLIP ViT-L/14 (\texttt{openai/clip-vit-large-patch14}) with CLIP's normalisation at $224\times224$, used only for the layer diagnostic of \cref{sec:where}.

\paragraph{Autoencoders.} SD ft-MSE (\texttt{stabilityai/sd-vae-ft-mse}; bit-identical to the autoencoder shipped with Stable Diffusion~2), SD1 original decoder (\texttt{stable-diffusion-v1-5}, \texttt{vae}), SDXL (\texttt{stabilityai/sdxl-vae}), Kandinsky~2.1 and 2.2 MoVQ, a FLUX-family autoencoder (\texttt{ostris/Flex.1-alpha}), and DC-AE (\texttt{mit-han-lab/dc-ae-f32c32-in-1.0}). The copy decodes the latent mode (no sampling); for VQ models, \texttt{decode} quantises. For Synthbuster we add OpenAI's consistency decoder (\texttt{openai/consistency-decoder}; two steps, fixed noise seed; run with fp16 weights because fp32 exceeded our GPU memory) and, post hoc, the autoencoder in the \texttt{vae\_1\_0} folder of \texttt{stabilityai/stable-diffusion-xl-base-1.0}. \texttt{stabilityai/sdxl-vae} is bit-identical to that repository's \texttt{vae} folder; the encoder and decoder weights of the two folders differ (largest absolute difference 0.016 and 0.042).

\paragraph{Views.} Centre crop of at most $256\times256$ on the 8-pixel grid (32-pixel grid for the autoencoder bank, so that DC-AE applies), copied at native resolution, then resized to the backbone's input with bicubic interpolation.

\paragraph{Test-time cost.} On the laptop used for all experiments (Apple M5, 16\,GB unified memory, PyTorch MPS backend), with $256$-pixel views of 64 GenImage images after one warm-up batch, per-image cost is 58\,ms for SimLBR and the supervised probe (one DINOv3 pass) and 58\,ms for \dvI{}, which reads the image alone and needs no copy at test time; its test-time path differs from SimLBR's only in the view (a native-resolution crop instead of a squashed image) and in the hidden states its head reads. \dvP{} and \dv-TF, which copy the image with ft-MSE and read image and copy, take 329\,ms; RIGID (two DINOv2 passes) 106\,ms; and AEROBLADE (three autoencoders and LPIPS) 699\,ms. Batches had 16 images, except for AEROBLADE, which exceeds the laptop's memory at that size and was timed with batches of 4 (\texttt{latency.py}).

\paragraph{Degraded real images.} Each test real receives a Gaussian blur with $\sigma\sim\mathcal U(0.5, 1.5)$ (seed 0, one draw per image) and then the same re-save as the other test images.

\section{RIGID: noise convention and protocol}
\label{app:rigid}
The RIGID paper specifies $\lambda=0.05$ and DINOv2 ViT-L/14 but not where the noise is added or how images are preprocessed. On a fixed GenImage subset of 4{,}000 images we compare noise added in $[0,1]$ pixel space with noise added to the normalised input, each with and without the JPEG q95 re-save. TPR@5\%: pixel noise 30.9 (raw) / 29.1 (q95); normalised noise 44.1 (raw) / 38.9 (q95). The JPEG re-save therefore costs RIGID at most 5 points and does not explain its weakness on latent diffusion images. Normalised noise is stronger on GenImage but flags 49.8\% of degraded real images without the re-save. \Cref{tab:main} reports both conventions on every evaluation image.

\section{All methods on the AEROBLADE subset}
\label{app:subset}
\Cref{tab:subset} scores every method on the same subset rows as AEROBLADE, so all comparisons with AEROBLADE are paired. On these rows the \dvP{} + SimLBR fusion beats the AEROBLADE + SimLBR fusion by 2.6 points \ci{0.8}{4.4} on GenImage and 2.5 \ci{0.5}{4.8} on Chameleon, and the AEROBLADE fusion flags 17--18\% of degraded real images. On Synthbuster, where the generated images were never compressed, AEROBLADE might be handicapped by the JPEG re-save that every evaluation image receives. Scored on the original files of the same rows, it detects 10.8\% at a 5\% false-positive rate (8.9\% with the re-save), and 7.0--8.5\% of SD~1.3, 1.4 and 2 images, whose autoencoders it uses.
\input{sections/tab_subset}

\section{Processing-aware training}
\label{app:procaware}
If processing is the problem, training on processed rows should fix it. We retrain \dv{} with every training row passed through the same random processing: JPEG at quality 60--90 with probability 0.5, then a resize from $\{\times0.5,\dots,\times1.5\}$ with probability 0.7, applied identically to the real, simulated-fake and counterfactual rows. On the controlled grid this recovers part of the loss: $\times1.5$ goes from 72.9\% to 85.3\%, and $\times0.75$ + q75 from 78.4\% to 83.3\%. It does not help on Chameleon (\dvP{} 48.1\% vs.\ 49.2\%), and it costs the untouched Chameleon fakes 17 points (75\% $\to$ 58\%). The false-positive rate on JPEG q75 reals at a fixed threshold stays at 12--16\%, and on resized reals it roughly doubles (\cref{fig:processing}b; $\times0.5$: 6.4\% $\to$ 11.2\%, $\times1.5$: 7.0\% $\to$ 12.8\%). Training on processed images makes the detector more sensitive to processing itself.

\section{Layer and counterfactual ablations}
\label{app:ablation}
\paragraph{Provenance of the final-layer variant.} The final-layer variant is the first version of \dv{}. Its heads were trained and scored on GenImage and Chameleon before the training-free layer diagnostic (\cref{sec:where}) led us to layers 8--18. We made that switch before the intermediate-layer heads had been evaluated; it raised GenImage and lowered Chameleon, and we kept it. Its Synthbuster scores, the processing and blur tests and the linear probes below were computed after all main results and are post hoc.

\paragraph{One factor at a time.} \Cref{tab:ablation} changes one factor of \dvI{} at a time; everything else (training images, head, seeds, evaluation) is unchanged. Reading the final layer instead of layers 8--18 costs 8.0 points on GenImage \ci{6.5}{9.4} but gains 27.2 on Chameleon \ci{24.4}{30.2}. For \dvP{} the same change costs 2.5 points on GenImage \ci{0.9}{4.1} and gains 14.9 on Chameleon \ci{12.5}{17.7}. On Synthbuster with downscaled reals (intervals resample photographs), the final layer costs \dvI{} nothing ($+1.1$ \ci{$-1.2$}{3.4}) and raises its worst generator by 34.8 points \ci{29.4}{40.3}; for \dvP{} it gains 12.2 \ci{9.8}{14.5}. With cropped reals the final layer gains 18.1 points for \dvI{} \ci{14.5}{21.4} and 20.7 for \dvP{} \ci{16.1}{25.1}, so the final-layer heads barely depend on the matching (\dvI{}: 85.7\% and 86.3\%; ahead of SimLBR by 5.5 \ci{2.9}{8.6} and 8.4 \ci{5.9}{10.9}). The change is not uniform across generators. With the final layer, \dvI{} detects far more GLIDE (69.7\% vs.\ 35.1\%) and DALL$\cdot$E~3 images (84.4\% vs.\ 62.6\%), and fewer images of SD~2, whose decoder makes the copies (86.7\% vs.\ 98.8\%), DALL$\cdot$E~2 (80.0\% vs.\ 93.7\%) and Firefly (88.7\% vs.\ 98.3\%). This is what one would expect if the intermediate layers carry most of the copy decoder's trace and the final layer a more generic difference that a trained head can also use; the training-free score, which has no head, detects 40.2\% of SD~2 images at layer 16 and 25.4\% at the final layer. Layers 8--18 are therefore not uniformly best for a trained head.

\paragraph{What the final layer reads.} The final layer is not blind to the copy. A linear head trained on the same rows as \dvI{} (real photographs, their copies and the counterfactual reals; same optimiser, steps and seeds) detects 73.2\% of GenImage's generated images from the final-layer class token, against 79.5\% for the MLP head, and 60.8\% and 90.4\% on Chameleon and Synthbuster with downscaled reals (MLP 61.2\% and 85.7\%; with cropped reals 88.8\% and 86.3\%). At layers 8--18 the linear head gives 85.6\%, 38.3\% and 85.6\% (MLP 87.5\%, 33.9\% and 84.5\%), and 52.4\% with cropped reals (MLP 68.2\%). To test the blindness reported by \citet{zhou2026simplicity} directly, we copied 1{,}000 GenImage test reals (long side at most 640 pixels) once at full resolution with ft-MSE, the SD~2 autoencoder, and scored the copies with every detector at its usual threshold. The originals are flagged at 4--5\%. Probes trained on SD~1.4 images flag 15--25\% of the copies (SimLBR released 24.9\%, retrained 20.1\%, supervised probe 15.2\%), close to the blindness Zhou et al.\ report. Heads trained on copies flag more: the reconstruction probe 32.0\% (it was trained on copies of squashed images, whereas here the image is squashed after the copy), final-layer \dvI{} 50.1\%, and \dvI{} on layers 8--18 97.6\%. Both the training target and the layer therefore matter.

The final-layer head is not robust to processing either. On the controlled grid of \cref{sec:wild} (SD~1.4, Wukong and Midjourney), it starts at 78.6\% (layers 8--18: 97.2\%) and is below the intermediate-layer head under every processing step. It loses less to resizing (a $\times1.5$ upscale: $+1.8$ points, against $-2.2$) and to the mild blur of \cref{sec:wild} (79.5\% $\to$ 64.0\% on the 1{,}600 GenImage fakes, against 87.9\% $\to$ 61.1\%), but as much or more to JPEG q75 ($-15.0$, against $-9.7$), and its false-alarm rate on processed reals at a fixed threshold stays between 2.2\% and 8.0\% (layers 8--18: 6.6--18.6\%). On Chameleon it detects as many generated images whose sides are not multiples of 8 pixels, which have almost certainly been resized, as untouched ones (61.6\% vs.\ 57.5\%); the intermediate-layer head detects 28.5\% vs.\ 51.1\%. The final-layer head thus reads a difference that is less tied to the copy decoder and less sensitive to resizing and blur, but not to compression. We have not identified what this difference is, and we do not claim that it survives web processing.

\paragraph{Counterfactual rows.} Removing the counterfactual rows raises TPR@5\% slightly (by 5.3 points on GenImage \ci{4.7}{6.0}, 1.9 on Chameleon \ci{0.9}{3.0} and 0.3 on Synthbuster \ci{$-0.6$}{1.4}) but multiplies false alarms on degraded real images by four to six. We did not retrain \dvI{} on squashed images; for the training-free signal, squashing the image to $256\times256$ before the copy costs 0.075 AUROC (\cref{sec:where}).

\begin{table}[h]
\caption{\textbf{\dvI{} with one factor changed.} The reference reads DINOv3 layers 8--18, copies at native resolution and trains with counterfactual rows. TPR (\%) at 5\% FPR (mean over generators) and false-positive rate (\%) on degraded real images at the same threshold; mean over 3 seeds. Synthbuster with downscaled reals. Paired intervals are in the text.}
\label{tab:ablation}
\centering\footnotesize\setlength{\tabcolsep}{5pt}
\begin{tabular}{@{}lrrrrrr@{}}
\toprule
 & \multicolumn{2}{c}{GenImage} & \multicolumn{2}{c}{Chameleon} & \multicolumn{2}{c}{Synthbuster} \\
\cmidrule(lr){2-3}\cmidrule(lr){4-5}\cmidrule(l){6-7}
Variant & TPR@5 & FPR$_\text{deg}$ & TPR@5 & FPR$_\text{deg}$ & TPR@5 & FPR$_\text{deg}$ \\
\midrule
\dvI{} as in the paper & 87.5 & 5.0 & 33.9 & 2.0 & 84.5 & 3.0 \\
\quad final layer instead of layers 8--18 & 79.5 & 3.6 & 61.2 & 3.9 & 85.7 & 4.0 \\
\quad without counterfactual rows & 92.8 & 29.8 & 36.0 & 12.8 & 84.9 & 16.8 \\
\bottomrule
\end{tabular}
\end{table}

\section{Synthbuster: per generator and further results}
\label{app:sbcrop}\label{app:sbmore}
\Cref{tab:sbcrop_gen} gives \dvI{} and the released SimLBR head per generator under the two ways of matching the real photographs (\cref{sec:synthbuster}), with paired intervals in which the calibration reals are resampled by photograph and each generator's generated images by image ($B=1{,}000$). Rows in which the difference has the same sign and excludes zero under both matchings are the ones the main text relies on. With cropped reals, training \dvI{} without counterfactual rows changes mean TPR@5\% by $-2.0$ \ci{$-5.3$}{0.8} (downscaled: $+0.3$ \ci{$-0.6$}{1.4}).

\paragraph{The paired variant and the training-free scores.} As on GenImage, \dvP{} is behind \dvI{} (74.7\%; $-9.8$ \ci{$-11.6$}{$-7.9$}; with cropped reals $-5.1$ \ci{$-9.3$}{$-1.4$}). Read at the final layer, \dvP{} detects 86.9\% and \dvI{} 85.7\% with downscaled reals (\cref{app:ablation}). The single-layer training-free scores fail after the JPEG re-save: \dv-TF detects 23.5\% (40.2\% even on SD~2, whose decoder is ours), RIGID with DINOv3 26.6\%, and RIGID with DINOv2, the published configuration, 5.4\%. AEROBLADE, on its fixed subset (\cref{app:subset}), detects 8.9\%, and at most 7\% of SD~1.3, SD~1.4 and SD~2 images although it uses their own autoencoders; \dvI{} is 77.1 points above it \ci{73.3}{80.7}. Without the re-save AEROBLADE detects 10.8\%, so the re-save is not why it fails here. The decoder bank of \cref{sec:decoder} removes the re-save and reads several layers.

\paragraph{Decoder bank: the other generators.} With downscaled reals, the consistency decoder's TPR on DALL$\cdot$E~3 (34.5\%) is the highest in its row, but its AUROC (0.75) is below those of ft-MSE, SDXL and FLUX (0.84--0.86). With cropped reals the prediction fails: the consistency decoder detects 38.0\% and the SDXL autoencoder 78.5\%. On the pixel-space generators no autoencoder exceeds 44\% (DALL$\cdot$E~2) and 41.5\% (GLIDE). Firefly is invisible to every autoencoder (at most 8.5\%), although \dvI{} detects 98\% of it. On Midjourney~v5, ft-MSE detects 91.0\% and attracts 81.5\% of the attributions. When each SD image is attributed to the autoencoder under which it moves least, 100\%, 98.5\% and 94.5\% of the SD~1.3, 1.4 and 2 images go to their own decoder, and with the post-hoc SDXL version 99.5\% of the SDXL images.

\begin{table}[h]
\caption{\textbf{Synthbuster per generator.} TPR (\%) at 5\% FPR of \dvI{} and SimLBR (released) with downscaled and with cropped reals, and their paired difference with a 95\% interval (calibration resampled by RAISE photograph). Mean over 3 seeds.}
\label{tab:sbcrop_gen}
\centering\footnotesize\setlength{\tabcolsep}{4pt}
\begin{tabular}{@{}lrrrrrr@{}}
\toprule
 & \multicolumn{3}{c}{Downscaled reals} & \multicolumn{3}{c}{Cropped reals} \\
\cmidrule(lr){2-4}\cmidrule(l){5-7}
Generator & \dvI{} & SimLBR & Difference & \dvI{} & SimLBR & Difference \\
\midrule
DALL$\cdot$E~2 & 93.7 & 50.7 & +43.0 [+38.5, +47.9] & 78.3 & 47.4 & +30.9 [+25.2, +36.0] \\
DALL$\cdot$E~3 & 62.6 & 99.8 & $-$37.2 [$-$42.0, $-$32.5] & 26.1 & 99.6 & $-$73.5 [$-$79.1, $-$66.5] \\
Firefly & 98.3 & 41.9 & +56.4 [+51.5, +63.2] & 90.5 & 36.1 & +54.4 [+49.2, +58.4] \\
GLIDE & 35.1 & 74.9 & $-$39.8 [$-$44.5, $-$33.3] & 13.1 & 71.3 & $-$58.2 [$-$62.1, $-$53.3] \\
Midjourney~v5 & 86.1 & 91.0 & $-$4.9 [$-$7.7, $-$1.5] & 60.2 & 90.0 & $-$29.8 [$-$36.7, $-$23.9] \\
SD~1.3 & 94.6 & 95.7 & $-$1.1 [$-$3.2, +1.2] & 81.9 & 94.5 & $-$12.6 [$-$16.3, $-$8.7] \\
SD~1.4 & 94.1 & 95.0 & $-$0.9 [$-$3.2, +2.0] & 82.3 & 93.2 & $-$10.9 [$-$15.6, $-$7.5] \\
SD~2 & 98.8 & 78.2 & +20.6 [+15.7, +26.1] & 93.7 & 73.8 & +20.0 [+16.2, +23.9] \\
SDXL & 97.6 & 95.9 & +1.7 [$-$0.1, +3.9] & 87.3 & 94.9 & $-$7.6 [$-$11.5, $-$4.0] \\
\bottomrule
\end{tabular}
\end{table}

\section{Decoder bank on GenImage and Chameleon}
\label{app:bankorig}
\begin{figure}[h]
    \centering
    \includegraphics[width=\linewidth]{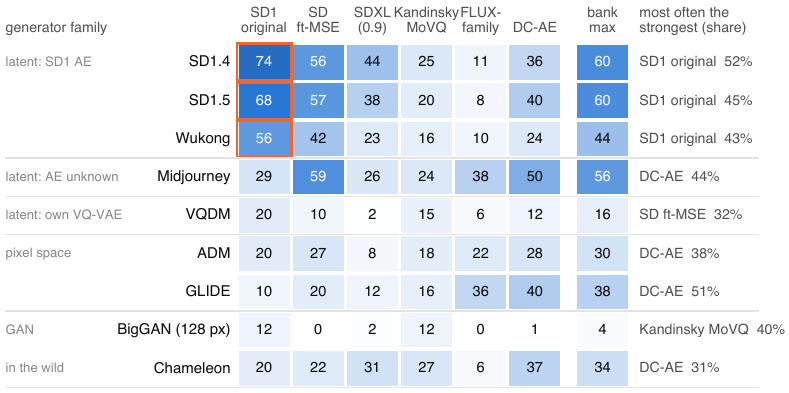}
    \caption{\textbf{Decoder bank on GenImage and Chameleon.} TPR (\%) at 5\% FPR of the training-free signal (DINOv3 class token, layers 10--18) per generator (rows) and autoencoder (columns). GenImage images are re-saved as JPEG q95 before the copy, so that real (JPEG) and generated (PNG) images have the same format; \cref{app:bankorig} shows the bank on the original files. The threshold is set in-sample, at the 95th percentile of the same set's real images (300 GenImage reals, 200 fakes per generator; 300 + 300 Chameleon). Orange outline: the generator's own decoder. Right: the autoencoder that most often gives a generator's images their strongest signal, and its share. With 200 fakes per cell, a 95\% binomial interval spans at most $\pm7$ points.}
    \label{fig:bank}
\end{figure}

\begin{figure}[h]
    \centering
    \includegraphics[width=0.92\linewidth]{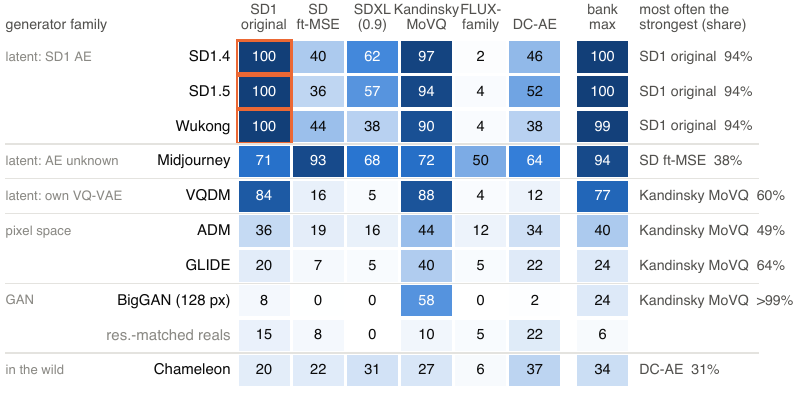}
    \caption{\textbf{Decoder bank on the original files} (JPEG real images, PNG generated images), for comparison with \cref{fig:bank}. The grey row re-thresholds BigGAN on real images downscaled to 128 pixels. Same threshold rule and samples as \cref{fig:bank}.}
    \label{fig:bankorig}
\end{figure}

\Cref{fig:bank} gives the training-free bank on GenImage, with every view re-saved as JPEG q95, and on Chameleon (\cref{sec:decoder}).
\Cref{fig:bankorig} repeats the bank of \cref{fig:bank} on the files as distributed, without the JPEG re-save. GenImage's real images are JPEG and its generated images PNG. On the original files every autoencoder appears to detect the SD1-based images, including the unrelated MoVQ (90--97\%), and SD1's decoder appears to detect VQDM (84\%); with the re-save these cross-detections disappear (MoVQ 16--25\% on SD1-based images, SD1's decoder 20\% on VQDM). BigGAN's apparent detection by MoVQ (58\%) also disappears with the re-save. We report the re-saved version in the main text.

\section{Copies from a different decoder}
\label{app:e8}
\Cref{tab:e8} compares \dv{} trained on copies from SD1's original decoder with the main version, whose copies come from ft-MSE; everything else is identical (training images, heads, seeds, evaluation). The rows marked in the ``Pred.'' column carry the signs we wrote down before training (\cref{sec:decoder}); all hold. Beyond the predictions, the copy decoder moves whole benchmarks, and not in one direction. On GenImage the SD1-decoder detector is better overall (\dvI{} +4.0 \ci{3.2}{4.9}; \dvP{} +11.4 \ci{10.5}{12.3}). This includes gains of about 11 points for \dvI{} on ADM, BigGAN and GLIDE, which have no SD1 decoder, so this gain is not decoder matching; we have not isolated its cause. On Synthbuster \dvI{} loses 6.0 points \ci{$-7.1$}{$-5.1$}, mostly on Midjourney ($-24.5$), DALL$\cdot$E~3 ($-18.3$) and SDXL ($-11.3$), while \dvP{} gains 3.4 \ci{2.2}{4.7}. On Chameleon both lose about 20 points.

\input{sections/tab_e8}

\section{Training-data scale for SimLBR}
\label{app:scale}
\begin{figure}[h]
    \centering
    \includegraphics[width=\linewidth]{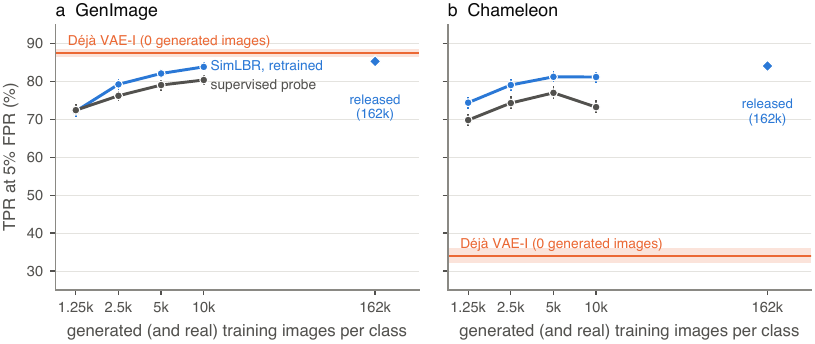}
    \caption{\textbf{Training-data scale of the baselines trained with generated images.} TPR at 5\% FPR of SimLBR retrained with its released recipe (blue) and of the plain supervised probe (grey) against the number of training images per class (nested subsets; equal numbers of real and SD~1.4 images), with the released SimLBR head (diamond, 162k per class). Orange line and band: \dvI{}, trained on 5{,}000 real images and no generated ones. Points: mean over 3 seeds; bars and band: 95\% bootstrap intervals (calibration reals and each generator's fakes resampled, $B=1{,}000$, seed-averaged).}
    \label{fig:scale}
\end{figure}

SimLBR retrained with nested subsets of 1{,}250, 2{,}500, 5{,}000 and 10{,}000 images per class reaches TPR@5\% of 72.1, 79.2, 82.1 and 83.8 on GenImage and 74.4, 79.0, 81.2 and 81.1 on Chameleon. The released head (162k per class) reaches 85.3 and 84.0. The plain supervised probe follows the same curve 2--8 points lower (\cref{fig:scale}). On GenImage, \dvI{} with no generated images (87.5) is above every retrained baseline; on Chameleon it is far below all of them.

\section{Size-matched Chameleon and the processing-only control}
\label{app:cham}
\begin{figure}[h]
    \centering
    \includegraphics[width=0.94\linewidth]{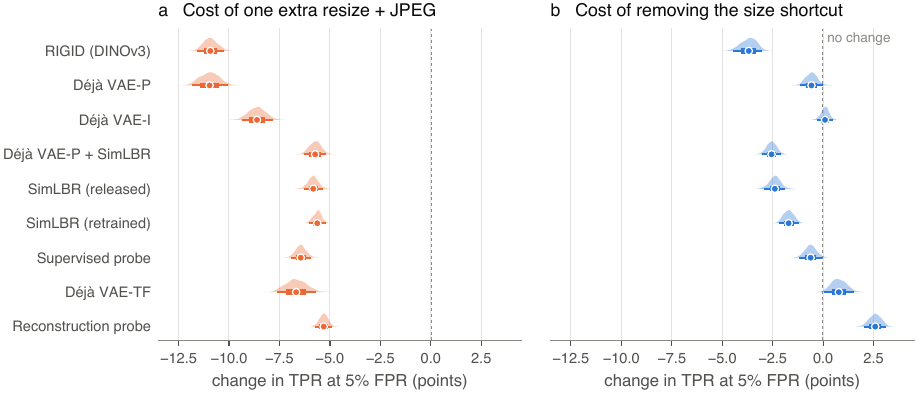}
    \caption{\textbf{Most of every detector's drop on size-matched Chameleon comes from processing, not from the shortcut.} Change in TPR@5\% caused by \textbf{(a)} one extra resize and JPEG q75 re-save (original $\to$ processing-only control) and \textbf{(b)} removing the size and aspect shortcut (control $\to$ size-matched). Slabs: paired bootstrap distributions (calibration reals and test fakes resampled jointly, $B=1{,}000$, mean over 3 seeds); dots: means; thick and thin bars: 66\% and 95\% intervals. Rows sorted by total change.}
    \label{fig:decomp}
\end{figure}

For each fake $i$ (seed 17) we draw a real image $j$ uniformly and set the target size to $(500, H_j)$. The fake is centre-cropped to aspect $H_j/500$ (height cropped if the fake is relatively taller, width otherwise), resized with bicubic interpolation (antialiased), and saved with PIL as baseline JPEG, quality 75, 4:2:0 subsampling, no metadata. The median resize factor is 0.71; 37.9\% of fakes are upscaled; the crop keeps a median of 66\% of the image area. The processing-only control applies the same resize factor to the uncropped fake (so its aspect ratio and full content are kept) and the same JPEG settings. Metadata audit: \cref{fig:audit}b; decomposition with intervals: \cref{fig:decomp}.

\section{Fusion against control fusions}
\label{app:fusion}
\begin{figure}[h]
    \centering
    \includegraphics[width=\linewidth]{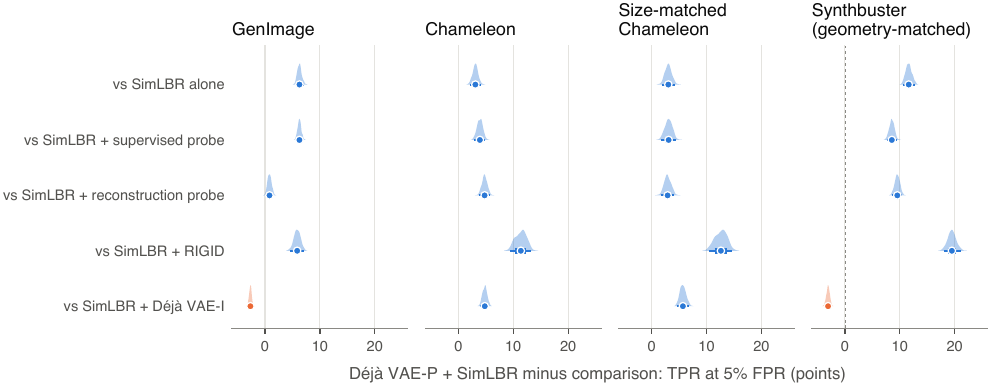}
    \caption{\textbf{Autoencoder traces add what a second semantic detector does not.} Paired differences in TPR@5\% between the \dvP{} + SimLBR fusion and, from top to bottom, SimLBR alone and four control fusions that replace \dvP{} with another detector, on each benchmark. Slabs: paired bootstrap distributions (calibration reals and test fakes resampled jointly, $B=1{,}000$, mean over 3 seeds); dots: means; thick and thin bars: 66\% and 95\% intervals. Blue: the fusion is better (95\% interval above zero); orange: worse. All panels share one scale.}
    \label{fig:fusion}
\end{figure}

\Cref{fig:fusion} shows the paired differences behind \cref{sec:fusion}.

\section{Reproducibility}
\label{app:repro}
Code and data: \url{https://github.com/iamsaugatpandey/deja-vae}. The repository contains:
\begin{itemize}
    \item the code for every experiment and figure, with the pipeline scripts in the order they were run;
    \item per-image scores of every method and seed on every benchmark, including degraded real images and the AEROBLADE subsets, keyed by each image's path relative to its dataset's root, so that every number in the paper can be recomputed without images;
    \item the lists that rebuild size-matched Chameleon, its processing-only control and both matched Synthbuster splits (downscaled and cropped) from the original datasets (file names and target sizes only, no image content);
    \item the decoder-bank reports for Synthbuster, as produced before and after the post-hoc SDXL analysis.
\end{itemize}
We release no images. GenImage, Chameleon, Synthbuster and RAISE-1k are available from their authors. All public model weights are loaded by name from their public repositories (\cref{app:impl}).

%% file: sections/tab_subset.tex
\begin{table}[h]
\caption{\textbf{All methods on the fixed stratified subsets} used for AEROBLADE (seed 13). GenImage: 1{,}000 calibration reals, 1{,}000 test reals stratified by source, 250 fakes per generator; Chameleon and size-matched Chameleon: 1{,}000 + 1{,}000 reals and 2{,}000 fakes; Synthbuster: 1{,}000 calibration reals, 1{,}800 geometry-matched test reals and 200 fakes per generator. TPR (\%) at 5\% and 1\% FPR, and FPR (\%) on degraded reals at the 5\% threshold.}
\label{tab:subset}
\centering\footnotesize\setlength{\tabcolsep}{2.8pt}
\begin{tabular}{@{}lrrrrrrrrrrrr@{}}
\toprule
 & \multicolumn{3}{c}{GenImage} & \multicolumn{3}{c}{Chameleon} & \multicolumn{3}{c}{Size-matched} & \multicolumn{3}{c}{Synthbuster}\\
\cmidrule(lr){2-4}\cmidrule(lr){5-7}\cmidrule(lr){8-10}\cmidrule(l){11-13}
Method & @5 & @1 & FPR$_\text{deg}$ & @5 & @1 & FPR$_\text{deg}$ & @5 & @1 & FPR$_\text{deg}$ & @5 & @1 & FPR$_\text{deg}$\\
\midrule
SimLBR (released) & 83.5 & 73.4 & 3.4 & 84.5 & 71.6 & 7.5 & 75.9 & 59.2 & 7.5 & 81.9 & 64.9 & 4.8\\
SimLBR (retrained) & 83.3 & 69.7 & 4.1 & 81.8 & 70.8 & 7.3 & 74.6 & 60.5 & 7.3 & 82.7 & 69.9 & 5.4\\
Supervised probe & 80.9 & 70.7 & 4.4 & 74.1 & 57.4 & 5.9 & 66.6 & 49.3 & 5.9 & 79.2 & 68.0 & 6.0\\
Reconstruction probe & 78.1 & 61.4 & 9.5 & 59.1 & 28.9 & 11.2 & 56.0 & 29.5 & 11.2 & 72.8 & 55.5 & 13.1\\
\dvI & 85.8 & 72.3 & 3.1 & 37.3 & 18.8 & 3.1 & 27.7 & 11.4 & 3.1 & 85.5 & 75.0 & 3.4\\
\dvP & 76.7 & 64.0 & 3.2 & 53.4 & 31.3 & 2.9 & 41.1 & 19.7 & 2.9 & 76.2 & 64.6 & 3.1\\
\dv-TF & 42.7 & 21.6 & 4.4 & 30.0 & 8.0 & 0.7 & 24.1 & 5.0 & 0.7 & 23.6 & 11.4 & 3.2\\
RIGID, DINOv2 (pixel) & 30.8 & 13.8 & 19.8 & 4.7 & 0.0 & 19.5 & 4.2 & 0.4 & 19.5 & 5.2 & 0.9 & 14.6\\
RIGID, DINOv2 (normalised) & 36.8 & 21.0 & 20.7 & 2.0 & 0.1 & 28.5 & 3.6 & 0.3 & 28.5 & 3.1 & 1.2 & 10.2\\
RIGID, DINOv3 & 34.2 & 17.2 & 17.2 & 27.4 & 7.4 & 15.6 & 13.4 & 2.1 & 15.6 & 26.7 & 11.8 & 16.6\\
AEROBLADE (min over 3 AEs) & 56.8 & 25.1 & 48.0 & 5.2 & 0.8 & 22.9 & 2.0 & 0.2 & 22.9 & 8.9 & 0.7 & 25.0\\
\quad AEROBLADE, SD1 & 58.2 & 32.2 & 49.1 & 4.1 & 0.4 & 29.7 & 1.8 & 0.1 & 29.7 & 6.8 & 0.9 & 21.8\\
\quad AEROBLADE, SD~2 (ft-MSE) & 41.5 & 18.8 & 47.8 & 5.2 & 0.8 & 22.7 & 2.0 & 0.2 & 22.7 & 8.4 & 0.9 & 25.3\\
\quad AEROBLADE, Kandinsky 2.1 & 32.6 & 14.9 & 51.4 & 3.8 & 0.4 & 27.4 & 1.8 & 0.1 & 27.4 & 5.3 & 0.6 & 21.8\\
\dvP{} + SimLBR & 89.7 & 80.6 & 2.4 & 87.8 & 74.2 & 3.5 & 78.2 & 60.3 & 3.5 & 92.1 & 85.6 & 3.0\\
AEROBLADE + SimLBR & 87.2 & 78.9 & 17.4 & 85.5 & 70.3 & 18.0 & 76.2 & 57.8 & 18.0 & 81.9 & 69.8 & 13.2\\
\bottomrule
\end{tabular}
\end{table}

%% file: sections/tab_e8.tex
\begin{table}[h]
\caption{\textbf{Copies from SD1's original decoder instead of ft-MSE.} Paired difference in TPR (\%) at 5\% FPR, SD1-decoder detector minus ft-MSE detector, with 95\% bootstrap intervals (calibration reals and each generator's fakes resampled jointly, seed-averaged, $B=1{,}000$). ``Pred.'': pre-specified sign (+ (P): predicted for \dvP{} only, as \dvI{} is at ceiling). ``ft-MSE'': \dvI{}'s TPR with ft-MSE copies.}
\label{tab:e8}
\centering\footnotesize\setlength{\tabcolsep}{4pt}
\begin{tabular}{@{}llrrrr@{}}
\toprule
Generator & Pred. & ft-MSE & $\Delta$ \dvI{} & $\Delta$ \dvP{} & $\Delta$ \dv-TF \\
\midrule
\multicolumn{6}{@{}l}{\textit{GenImage}} \\
ADM &  & 66.8 & +11.3 {\scriptsize[+9.3, +13.5]} & +22.4 {\scriptsize[+19.8, +25.0]} & $-$3.6 {\scriptsize[$-$6.6, +0.0]} \\
BigGAN &  & 83.8 & +10.8 {\scriptsize[+9.3, +12.6]} & +20.4 {\scriptsize[+18.6, +22.5]} & +11.4 {\scriptsize[+9.9, +13.8]} \\
Midjourney & $-$ & 91.3 & $-$7.7 {\scriptsize[$-$9.5, $-$6.0]} & $-$2.7 {\scriptsize[$-$3.9, $-$1.3]} & $-$33.1 {\scriptsize[$-$35.8, $-$29.4]} \\
VQDM &  & 95.9 & +3.8 {\scriptsize[+2.9, +4.8]} & +16.6 {\scriptsize[+14.7, +18.6]} & +9.1 {\scriptsize[+6.4, +12.4]} \\
GLIDE &  & 65.1 & +11.5 {\scriptsize[+9.1, +14.1]} & +30.4 {\scriptsize[+27.7, +33.3]} & $-$10.8 {\scriptsize[$-$14.0, $-$7.1]} \\
SD~1.4 & + (P) & 99.5 & +0.4 {\scriptsize[+0.2, +0.8]} & +0.8 {\scriptsize[+0.4, +1.2]} & +7.9 {\scriptsize[+4.6, +11.2]} \\
SD~1.5 & + (P) & 99.1 & +0.4 {\scriptsize[+0.0, +0.9]} & +1.3 {\scriptsize[+0.8, +1.8]} & +7.5 {\scriptsize[+4.9, +10.7]} \\
Wukong & + (P) & 98.7 & +1.2 {\scriptsize[+0.7, +1.7]} & +1.6 {\scriptsize[+1.1, +2.2]} & +7.8 {\scriptsize[+4.8, +11.4]} \\
\textit{mean} &  &  & +4.0 {\scriptsize[+3.2, +4.9]} & +11.4 {\scriptsize[+10.5, +12.3]} & $-$0.2 {\scriptsize[$-$2.3, +1.8]} \\
\multicolumn{6}{@{}l}{\textit{Synthbuster}} \\
DALL$\cdot$E 2 &  & 93.7 & $-$1.9 {\scriptsize[$-$3.4, $-$0.6]} & +3.4 {\scriptsize[+0.9, +6.2]} & $-$1.7 {\scriptsize[$-$4.0, +0.2]} \\
DALL$\cdot$E 3 &  & 62.6 & $-$18.3 {\scriptsize[$-$21.0, $-$15.0]} & +1.3 {\scriptsize[$-$1.9, +5.0]} & $-$6.8 {\scriptsize[$-$10.4, $-$3.2]} \\
Firefly &  & 98.3 & +0.3 {\scriptsize[$-$0.2, +1.0]} & +11.9 {\scriptsize[+9.8, +14.6]} & +5.7 {\scriptsize[+4.2, +7.4]} \\
GLIDE &  & 35.1 & $-$6.7 {\scriptsize[$-$9.6, $-$4.0]} & +11.9 {\scriptsize[+9.2, +14.5]} & $-$13.3 {\scriptsize[$-$16.5, $-$10.4]} \\
Midjourney v5 &  & 86.1 & $-$24.5 {\scriptsize[$-$27.9, $-$21.7]} & $-$10.9 {\scriptsize[$-$13.6, $-$8.5]} & $-$28.9 {\scriptsize[$-$32.3, $-$25.1]} \\
SD~1.3 & + & 94.6 & +5.1 {\scriptsize[+3.9, +6.5]} & +7.6 {\scriptsize[+6.1, +9.2]} & +36.2 {\scriptsize[+32.1, +40.7]} \\
SD~1.4 & + & 94.1 & +5.3 {\scriptsize[+3.9, +6.6]} & +6.6 {\scriptsize[+5.1, +8.3]} & +35.9 {\scriptsize[+31.7, +40.7]} \\
SD~2 & $-$ & 98.8 & $-$2.1 {\scriptsize[$-$3.0, $-$1.1]} & $-$2.3 {\scriptsize[$-$3.3, $-$1.3]} & $-$32.8 {\scriptsize[$-$36.1, $-$29.5]} \\
SDXL &  & 97.6 & $-$11.3 {\scriptsize[$-$13.3, $-$9.2]} & +0.8 {\scriptsize[$-$2.0, +3.2]} & $-$0.3 {\scriptsize[$-$1.0, +0.5]} \\
\textit{mean} &  &  & $-$6.0 {\scriptsize[$-$7.1, $-$5.1]} & +3.4 {\scriptsize[+2.2, +4.7]} & $-$0.6 {\scriptsize[$-$2.2, +1.1]} \\
\multicolumn{6}{@{}l}{\textit{Chameleon}} \\
all fakes &  & 33.9 & $-$18.0 {\scriptsize[$-$19.7, $-$16.5]} & $-$20.8 {\scriptsize[$-$22.8, $-$18.6]} & $-$9.2 {\scriptsize[$-$11.4, $-$6.0]} \\
\bottomrule
\end{tabular}
\end{table}